\documentclass[journal]{IEEEtran}
\usepackage{flushend}  

\usepackage{cite}

\ifCLASSINFOpdf
  \usepackage[pdftex]{graphicx}
  \DeclareGraphicsExtensions{.pdf,.jpeg,.png}
\else
  \usepackage[dvips]{graphicx}
  \DeclareGraphicsExtensions{.eps}
\fi
\usepackage[cmex10]{amsmath}
\usepackage{algpseudocode}
\usepackage[]{algorithm}

\usepackage[caption=false,font=footnotesize]{subfig}
\newtheorem{prop}{Proposition}
\newtheorem{assum}{Assumption}

\newtheorem{defi}{Definition}
\newtheorem{lem}{Lemma}
\newtheorem{cor}{Corollary}

\usepackage{color}

\usepackage{multirow}

\usepackage{amsfonts}
\newcommand{\fr}[1]{\mathsf{#1}}
\newcommand{\vect}[1]{\mathbf{#1}}

\begin{document}
%
\title{A Motion Planning Strategy for the Active
Vision-Based Mapping of Ground-Level Structures}
%
%
%

\author{Manikandasriram~S.R.$^{1}$, 
Andr\'e~Phu-Van~Nguyen$^{2}$,
and Jerome~Le~Ny$^{2}$,~\IEEEmembership{Senior Member,~IEEE}
\thanks{Part of this work was performed while the first author was visiting 
Polytechnique Montreal, under a Globalink Fellowship from MITACS.
This work was also supported by NSERC (Grant 435905-13) and the 
Canada Foundation for Innovation (Grant 32848).
}
\thanks{$^{1}$M.~S.R. is with the Robotics Institute, University of Michigan, 
Ann Arbor, MI 48109, USA. {\tt\small srmani@umich.edu}}
\thanks{$^{2}$A. Phu-Van Nguyen and J. Le Ny are with the Department of Electrical Engineering, 
Polytechnique Montreal, and GERAD, Montreal, QC H3T 1J4, Canada. {\tt\small \{andre-phu-van.nguyen,jerome.le-ny\}@polymtl.ca}}
\thanks{Digital Object Identifier 10.1109/TASE.2017.2762088}
}

%
%

\markboth{IEEE Transactions on Automation Science and Engineering}{Srinivasan Ramanagopal \MakeLowercase{\textit{et al.}}: Motion Planning Strategy for the Active Vision-Based Mapping}
%

\IEEEpubid{\begin{minipage}{\textwidth}\ \\[12pt]
\centering
  1545--5955~\copyright~2017 IEEE. Personal use of this material is permitted.\\ Permission from IEEE must be obtained for all other uses, in any current or future media, including reprinting/republishing this material for advertising or promotional purposes, creating new collective works, for resale or redistribution to servers or lists, or reuse of any copyrighted component of this work in other works
\end{minipage}}


\maketitle


\begin{abstract}
This paper presents a strategy to guide a mobile ground robot equipped 
with a camera or depth sensor, in order to autonomously map the 
visible part of a bounded three-dimensional structure.
We describe motion planning algorithms that determine appropriate successive 
viewpoints and attempt to fill holes automatically in a point cloud 
produced by the sensing and perception layer. 
The emphasis is on accurately reconstructing a 3D model of a structure of 
moderate size rather than mapping large open environments, with 
applications for example in architecture, construction and inspection. 
The proposed algorithms do not require any initialization in the 
form of a mesh model or a bounding box, and the paths generated 
are well adapted to situations where the vision sensor 
is used simultaneously for mapping and for localizing the robot, in
the absence of additional absolute positioning system.
We analyze the coverage properties of our policy, and 
compare its performance to the classic frontier based exploration algorithm.
We illustrate its efficacy for different structure sizes, 
levels of localization accuracy and range of the depth sensor, 
and validate our design on a real-world experiment.

\emph{Note to Practitioners---}
The objective of this work is to automate the process of building a 3D model 
of a structure of interest that is as complete as possible, using a mobile 
camera or depth sensor, in the absence of any prior information about this structure.
Given that increasingly robust solutions for the Visual Simultaneous Localization 
and Mapping problem (vSLAM) are now readily available, the key challenge that 
we address here is to develop motion planning policies to control the 
trajectory of the sensor in a way that improves the mapping performance. 
We target in particular scenarios were no external absolute positioning system 
is available, such as mapping certain indoor environments where GPS signals are
blocked. In this case, it is often important to revisit previously seen locations
relatively quickly, in order to avoid excessive drift in the dead-reckoning localization
system.
Our system works by first determining the boundaries of the structure, 
before attempting to fill the holes in the constructed model. 
Its performance is illustrated through simulations and a real-world experiment
performed with a depth sensor carried by a mobile manipulator.
\end{abstract}



%



\section{Introduction}
%
%
%
%
%



\IEEEPARstart{A}{ccurate} 3D computer models of large structures have
a wide range of practical applications, from inspecting an aging structure 
to providing virtual tours of cultural heritage sites  
\cite{Jacobi201580,el2004detailed}. 
In civil engineering for example, an important problem is 
that of \emph{construction progress monitoring}, i.e., comparing 
the state of a building under construction over time to the project
plan. The process of regularly updating the estimate of the state of 
the building has traditionally been performed manually, but in recent 
years new methods have been developed to automate it using data 
obtained from a variety of sensors, e.g., positioning systems,
stationary 3D laser scanners \cite{Yelda:2012:3Dconstruction}, 
high resolution video cameras \cite{Brilakis:2011:videogrammetry},
or still cameras carried by UAVs \cite{Lin:civEng2015:UAVconstructionMonitoring}.
\IEEEpubidadjcol

This paper considers the problem of guiding in real-time a mobile autonomous 
robot carrying a vision sensor, in order to build a 3D model of a structure. For this,
we need to address two problems. 
First, we need a robust mapping system that can build the 3D model in real-time when 
given a sequence of images or depth maps as input. 
This is a widely researched problem called Visual Simultaneous Localization 
and Mapping (vSLAM) or real-time Structure from Motion (SfM), for which several open source 
packages offer increasingly accurate and efficient solutions 
\cite{labbe14online,Endres:TRO2014:RGBDSLAM}.
The second problem relates to active sensing \cite{Bajcsy:IEEE88:activePerception},
as we need motion planning strategies that can guide a mobile sensor to explore 
the structure of interest. For mapping, monitoring or inspection applications, 
certain classical strategies such as frontier-based exploration 
algorithms \cite{Yamauchi:1997:FAA:523996.793157}, which guide the robot to previously 
unexplored regions irrespective of whether it is part of the structure 
of interest or not, are not necessarily well adapted.



Some recent work considers the problem of reconstructing a 3D model
of arbitrary objects by moving a depth sensor relative to the object
using different forms of \emph{next best view} planning algorithms
\cite{kriegel2013efficient,krainin2011autonomous}. 
Typically, these systems iteratively build a complete 3D model of 
the object by heuristically choosing the next best viewpoint 
according to some performance measure.
However, much of this work is 
restricted to building models of relatively small objects that 
are bounded by the size of the robot workspace.
In contrast, our focus is on 3D reconstruction of larger but still
bounded structures such as buildings, which can be several orders 
of magnitude larger than a mobile robot. 
The related problem of automated inspection deals with large 
structures such as tall buildings \cite{Lin:civEng2015:UAVconstructionMonitoring}
and ship hulls.
Bircher et al. \cite{bircher2015structural} assume that
a prior 3D mesh of the structure to inspect is available and
compute a short path connecting viewpoints that together are guaranteed
to cover all triangles in the mesh.
In \cite{englot2012sampling}, Englot et al. begin by assuming
a safe bounding box of the hull and construct a coarse mesh
of the hull by tracing along the walls of this box in a
fixed trajectory without taking feedback from the actual 
geometry of the structure. Moreover, this coarse mesh is 
manually processed offline to yield an accurate 3D mesh, 
which is then used to inspect the finer structural details. 
Yoder and Scherer \cite{Yoder:FSR15:frontierMAV}
also assume a bounding box and develop an algorithm combining next best
view planning and frontier-based exploration to encourage coverage of the 
structure.
Sheng et al. \cite{Sheng08CrawlerInspection} use 
a prior CAD model of an aircraft to plan a path for a robotic crawler
such that it inspects all the rivets on the surface of the aircraft.
In this paper however, we do not assume any prior information 
in terms of a 3D mesh, CAD model or a bounding box around the
structure, and focus on reactive path-planning to build the
model online.
Our mapping problem is also related to coverage path planning, see, e.g., 
\cite{Hert:AR96:coverage, Acar:IJRR02:MoreCoverageControl, acar:IJRR02:coverage, 
Lim14CrackInspection} and the references therein, 
which has traditionally focused on developing algorithms ensuring that
a mobile robot passes over all points in a 2D environment,
assuming 
a sufficiently accurate localization system.

\begin{figure}[!tb]
\centering
\subfloat[]{
    \includegraphics[width=2.6in]{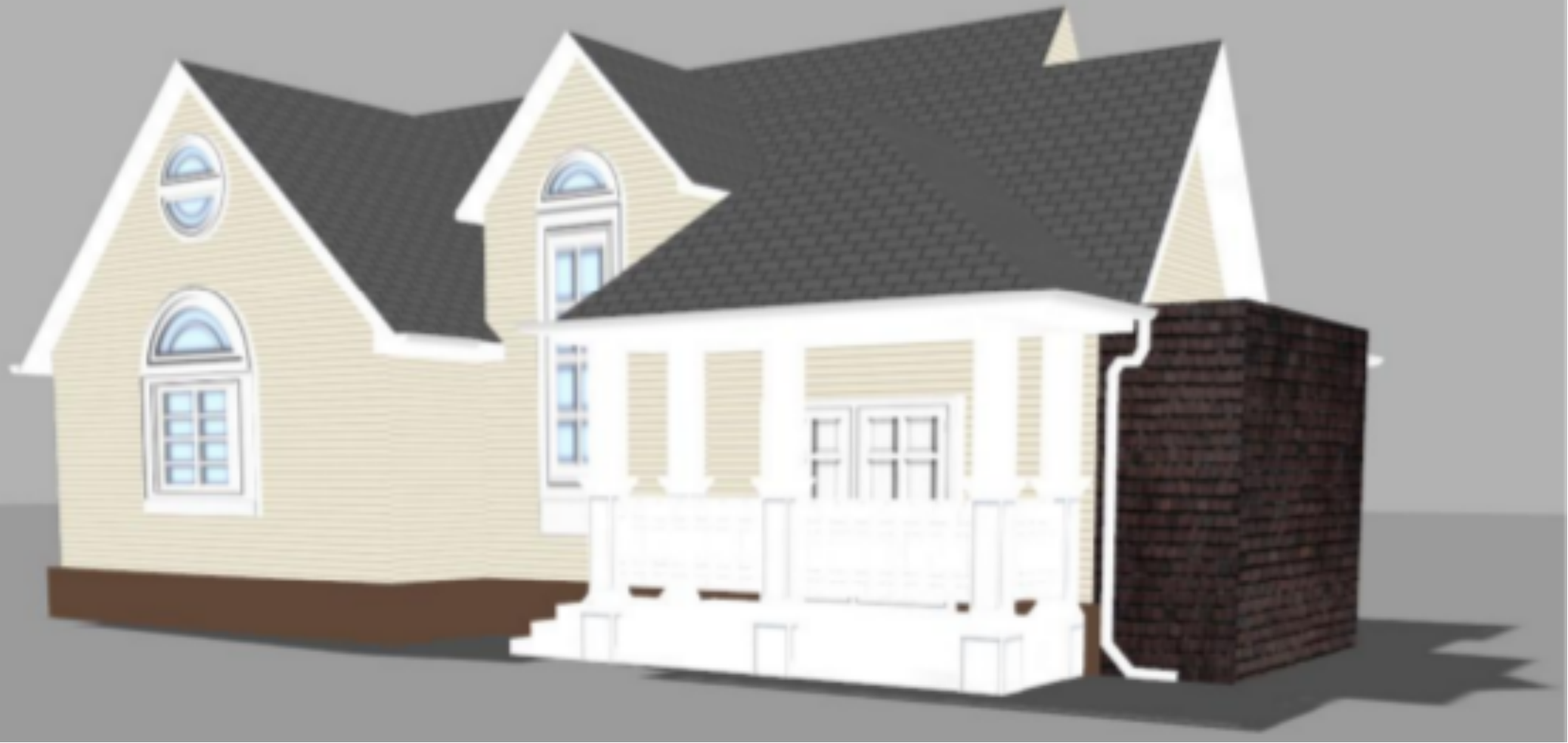}
	\label{subfig:gazebo_model}
}
\hfil
\subfloat[]{
    \includegraphics[width=2.6in]{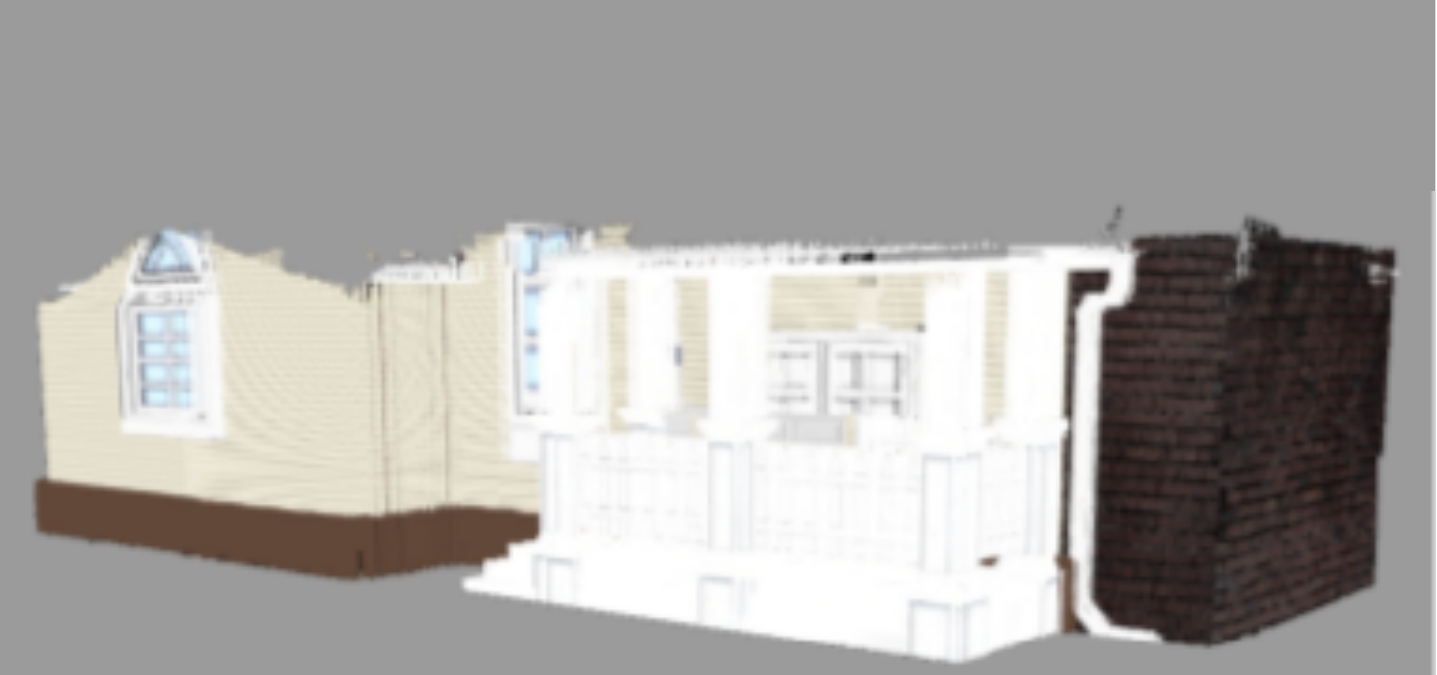}
	\label{subfig:reconstructed_model}
}
\caption{Comparison of the a) Simulated Model in Gazebo \cite{gazebo} that needs to be mapped and b) Reconstructed 3D model by a mobile ground robot using our policies. Only the bottom portion is mapped due to the limited reachable space of the sensor.}
\label{fig:comparison}
\end{figure}

In computer vision and photogrammetry, SfM techniques aim at building 
a 3D model of a scene from a large number of images 
\cite{Snavely:JCV08:SfM,frahm2010building,wu2011visualsfm,Furukawa:PAMI10:stereopsis},
but most of this work focuses on batch post-processing and typically 
assumes a given dataset, whereas here our focus is essentially on how
to acquire an appropriate set of images.
Let us mention however the work of Daftry et al. \cite{Daftry:ICRA15:UAVmapping},
which presents an interactive real-time SfM system providing online feedback to the user
taking pictures, alerting him or her when a new picture cannot be properly
integrated in the model. Also, Tuite et al. \cite{Tuite:SIGCHI11:SfM} develop
a competitive game where players are encouraged to take pictures that help build
complete 3D models.
We emphasize that 
we do not discuss in details the task of actually building a model 
from a collection of pictures or depth maps, which can be executed by one of the 
available vSLAM or real-time SfM systems, such as the Real-Time Appearance Based Mapping 
package (RTAB-Map) \cite{labbe14online} that we use in our experiments. 
Our work focuses on actively exploring the environment with an autonomous robot
to build a complete model in real-time, with our controller taking 
at any time the current model as an input.
Which package we use for model reconstruction has little influence on
our algorithms, for example any vSLAM system based on pose-graph 
optimization \cite{kuemmerle11icra} could be used. 
State-of-the-art batch SfM systems can also be used to post-process 
the sequence of images or depth maps captured using our policies in order 
to obtain a more accurate model offline. 
Naturally, eventual completeness of the model is limited by the 
physical characteristics of the robot, and specifically 
the reachable space of the sensor, see Fig. \ref{fig:comparison}.

Finally, another line of work in informative path planning relates to 
autonomous exploration and coverage of relatively large environments, using 
variants of frontier based exploration algorithms for example
\cite{shade2011choosing, shen2012autonomous, heng2015efficient, Atanasov:ICRA2015:activeSLAM}. 
While these papers focus on path planning to quickly build models of potentially large and complex spaces, 
they do not address the problem of autonomously delimiting and mapping 
as completely as possible a specific bounded structure of interest.

Our contributions can be summarized as follows. After presenting the problem statement
in Section \ref{section: statement}, we develop in Section \ref{section: PE} a motion 
planner allowing a ground robot equipped with a camera or depth sensor to autonomously 
determine the boundaries of an initially unknown structure. 
Then, Section \ref{section: CE} describes an algorithm for detecting missing portions 
in the 3D model constructed during the boundary determination phase, and an exploration 
strategy to improve the completeness of the model. 
In Section \ref{section: analysis} we analyze the level of coverage completeness that 
can be expected from our strategy.
The behavior of the proposed policies is illustrated in Section \ref{section: validation} 
through simulations, and the resulting accuracy of the constructed models compared to that
obtained using the classical frontier based exploration algorithm.
Experimental results are presented in Section \ref{section: experiments} to validate
the algorithms under more realistic illumination conditions.
One justification for our incremental exploration approach is that 
we focus on using the vSLAM module both for mapping the structure as well as localizing
the robot, although an additional dead-reckoning system such as wheel odometry could
be present as well. In the absence of an independent source of accurate absolute positioning,
it is important to close loops relatively frequently with the vSLAM system, i.e., revisit 
regions that have already been explored, in order to control the growth of the localization
errors building up with visual odometry alone. %
We also help the vSLAM system by following the boundaries of the structure, where visual
features are likely to be present.



\section{Problem Statement and Assumptions} 
\label{section: statement}

Consider the problem of constructing a 3D model of a given structure 
of finite size, e.g., a monument or a building, 
using a mobile ground robot carrying an imaging or depth sensor, 
such as a Kinect, 
a monocular or stereo camera, or a LIDAR.
Initially, no approximate model of the structure nor map of the environment 
is available, and the actual size of the structure is also unknown. 
The sensor (also called camera in the following) provides a sequence of point clouds 
obtained directly or computed from depth and/or luminance images. 
These local point clouds, together with an estimate of the sensor trajectory, 
can then be assembled and registered in a coordinate frame in real-time 
using available SLAM algorithms, such as RTAB-Map \cite{labbe14online} 
or RGBD-SLAM \cite{Endres:TRO2014:RGBDSLAM}, and post-processing then allows us to 
build a dense 3D model or a 3D occupancy grid stored in an OctoMap \cite{hornung13auro}.
We do not directly address here the model reconstruction problem in vSLAM. 
Instead, we focus on determining good trajectories for the robot allowing a vSLAM module
(and potentially a batch SfM module in post-processing) to produce a high quality model,
which ideally should capture the entire visible portion of the structure accurately.
A key challenge is to develop strategies that are applicable for any 
type of structure while respecting the physical limitations of the platform.

\begin{figure}[!t]
\centering
\subfloat[]{
	\includegraphics[width=1.62in]{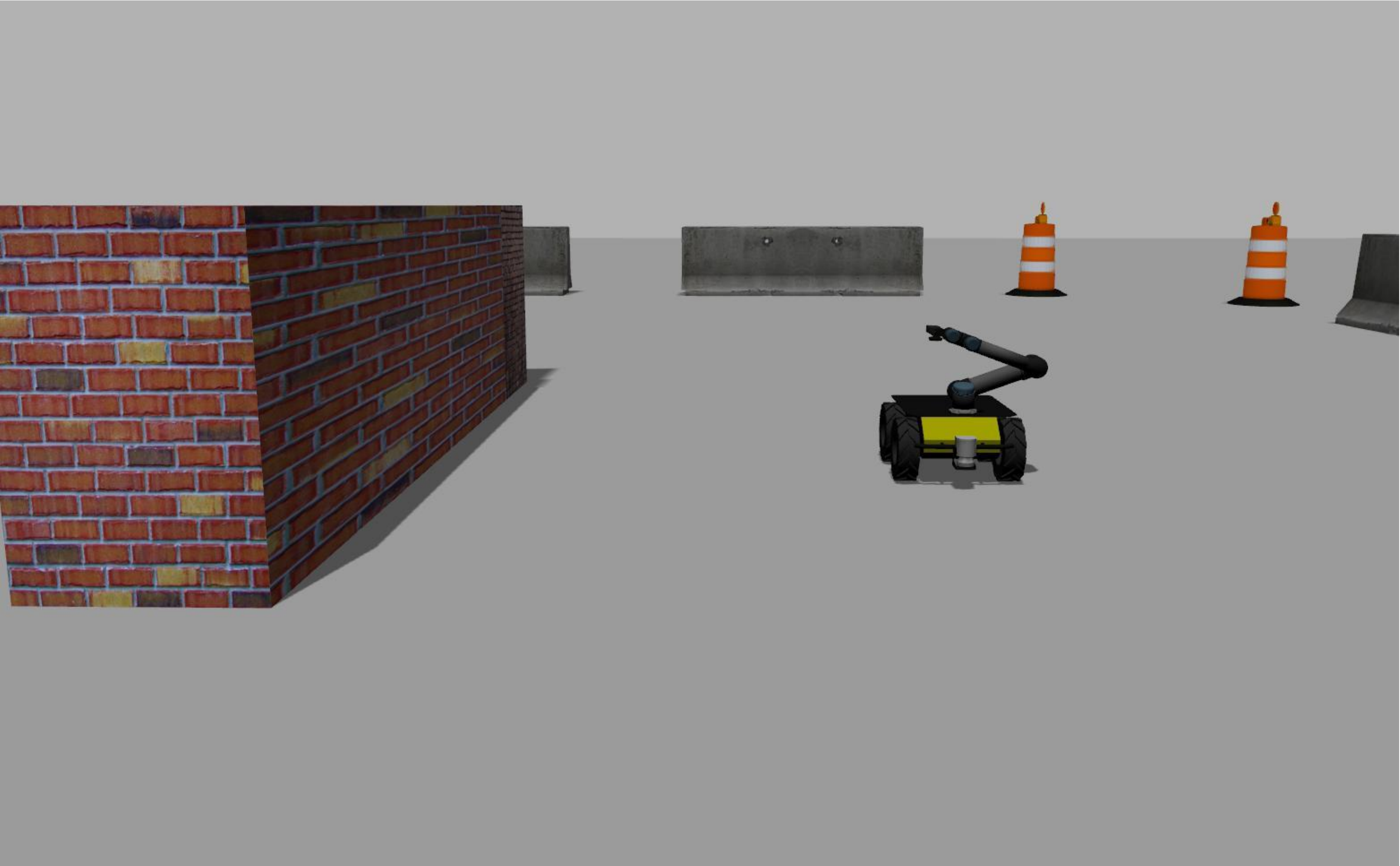}
	\label{subfig:initial conditions}
}
\hfil
\subfloat[]{
	\includegraphics[width=1.62in]{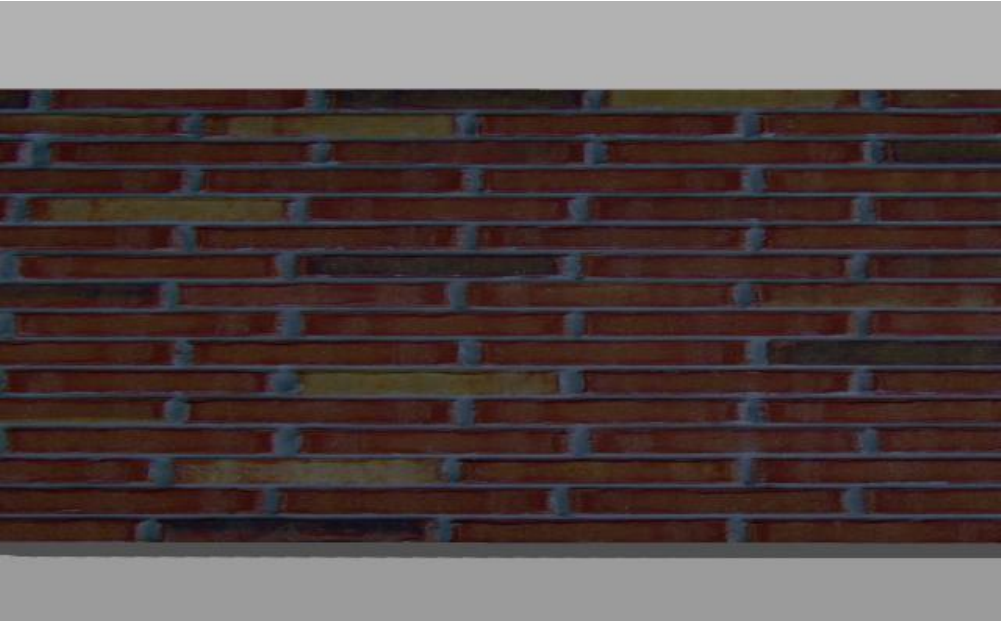}
	\label{subfig:initial image}
}
\caption{(a) 
Starting configuration for the robot and camera with respect to the structure.
(b) Initial image seen by the camera: the robot only knows that the structure 
in the field of view (FOV) is the one that should be mapped.
}
\label{fig:initial conditions}
\end{figure}

We assume that initially the robot is positioned along the structure to
be mapped, with the camera capturing point clouds mounted on its right and 
facing the structure at a distance $D$ measured in a horizontal plane, 
see Fig. \ref{fig:initial conditions}. 
In normal operations, we wish to maintain this distance $D$ between the structure
and the path of the camera, where $D$ is chosen based on the camera's resolution.
Define a global fixed Frame of Reference (FoR) 
$\fr G := \{O_{\fr g}, \vect{x}_{\fr g}, \vect{y}_{\fr g}, \vect{z}_{\fr g}\}$,
in which the global point cloud is to be assembled.
Note that we write vectors in bold.
The robot FoR $\fr R := \{O_{\fr r}, \vect{x}_{\fr r}, \vect{y}_{\fr r}, \vect{z}_{\fr r}\}$ 
(forward, left, up) coincides initially with $\fr G$, but is attached to a point 
$O_{\fr r}$ that moves along with the robot.
For concreteness to describe our scenario and algorithms,
the camera FoR $\fr C := \{O_{\fr c}, \vect{x}_{\fr c}, \vect{y}_{\fr c}, \vect{z}_{\fr c}\}$,
is assumed to be rigidly attached to the robot except for the yaw motion, 
which is left unconstrained. 
\begin{assum}\label{assumption: fixed camera}
The center $O_{\fr c}$ of the camera mounted on the mobile ground robot 
has fixed coordinates $(0,0,h_{\fr c})$ in frame $\fr R$,
and in addition we always maintain $\vect{z}_{\fr r} = \vect{z}_{\fr c}$.
\end{assum}
Such a choice of camera configuration determines which parts of the structure
are not visible at all, and hence cannot be mapped by any algorithm implemented 
on this platform. However, other system configurations could be handled with some of 
the more generic tools developed in this paper.

\begin{figure}[!ht]
	\centering
	\includegraphics[width=0.8\linewidth]{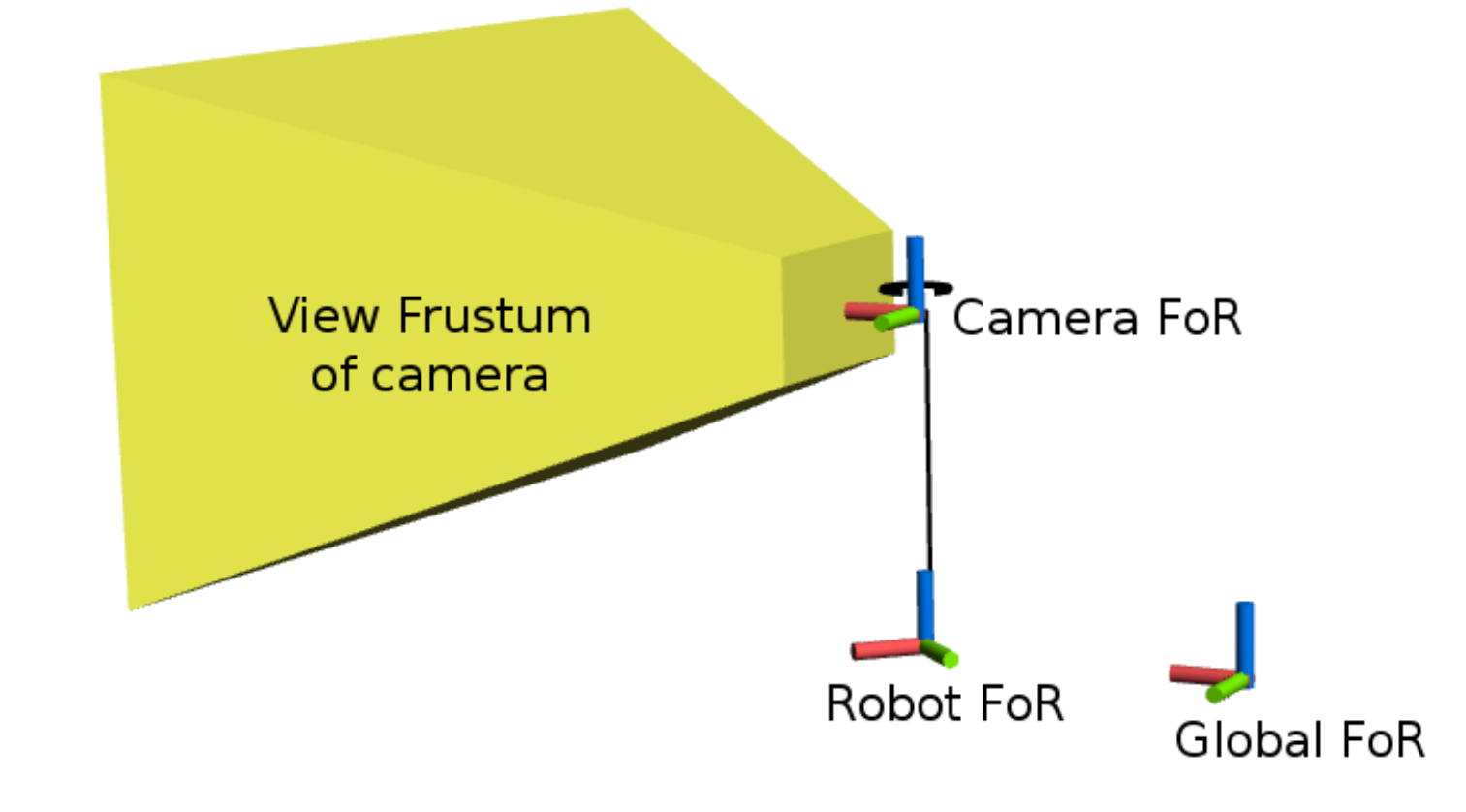}
	\caption{The camera is kept at a constant height above the robot's base. 
    The red, green and blue lines correspond to the $x$, $y$ and $z$ axes respectively 
    and both the camera and robot can rotate along their $z$ axes. 
    The yellow region corresponds to the view frustum of the camera. 
}
\label{fig:assumptions}
\end{figure}

Fig. \ref{fig:assumptions} shows our conventions for the different FoR used.
The imaging plane of the camera is defined by $\vect{y}_{\fr c} \vect{z}_{\fr c}$, with
$\vect{x}_{\fr c}$ pointing towards the front of the camera on the optical axis.
Coordinates in the camera, robot and global FoR are denoted using superscripts as 
$\vect{v}^{\fr c}$, $\vect{v}^{\fr r}$ and $\vect{v}^{\fr g}$ respectively for a vector $\vect{v}$.
We assume that initially $\mathbf x_\fr c = -\mathbf y_\fr r$ so that the camera
points to the right of the robot. 
We make two additional assumptions for simplicity of exposition.
The first one guarantees that there exists collision free paths around the structure.
\begin{assum}	\label{assumption: annulus} 
The horizontal distance of the closest obstacle from the structure is at least $2 D$.
\end{assum}
The next assumption simplifies the problem of detecting, tracking and removing the 
ground surface from point clouds, 
a processing step performed in Algorithm \ref{alg:gpp_computeNBV} to 
compute waypoints that only depend on the structure to inspect.
\begin{assum}\label{assumption: level ground}
The structure and the robot are placed on the horizontal plane 
$z^{\fr g} = 0$. 
In particular, we have $\vect{z}_{\fr c}=\vect{z}_{\fr r} = \vect{z}_{\fr g}$.
\end{assum}
\vspace{0.1cm}
In the following we fix the $z$-coordinate of $O_{\mathsf r}$ to be zero.
A consequence of these assumptions is that relatively horizontal surfaces that
are at the same height or above the camera center for example cannot be mapped, 
and the maximum height (measured in the $\fr R$ or $\fr G$ frame) of the structure 
that can be mapped is $H_{\max} = h_{\fr c} + D \tan{\psi/2}$,
where $\psi$ is the vertical angle of view of the camera. 
Assumption \ref{assumption: level ground} could be removed by using recent 
classification systems that can differentiate between ground and non-ground 
regions \cite{zhou2012self} to pre-process the point clouds before sending 
them to our system.

Finally, there are additional implicit assumptions that we state informally.
First, since we rely on an external mapping module to build the 3D model, 
the conditions that allow this module to operate sufficiently reliably must 
be met. For example, vSLAM generally requires appropriate scene 
illumination and the presence of a sufficiently rich set of visual features.
Second, we concentrate on the reconstruction of the details of the model 
at a scale comparable with or larger than the typical length of the robot.
If features at a smaller scale need to be included, e.g., fine structural
details on a wall, our system could be augmented with a more local planner
for a robotic arm carrying the sensor \cite{kriegel2013efficient, Dornhege13advro},
as well as targeted computer vision techniques \cite{Furukawa:PAMI10:stereopsis}.
Finally, for reasons explained in Section \ref{section: obstacle ahead}, we assume 
that the robot is equipped with sensors capable of detecting obstacles 
in a $180^{\circ}$ region ahead of it and within a distance of $D$, 
see Fig. \ref{fig:concave_surfaces}.

\begin{figure}[!htb]
\centering
\includegraphics[width=0.9\linewidth]{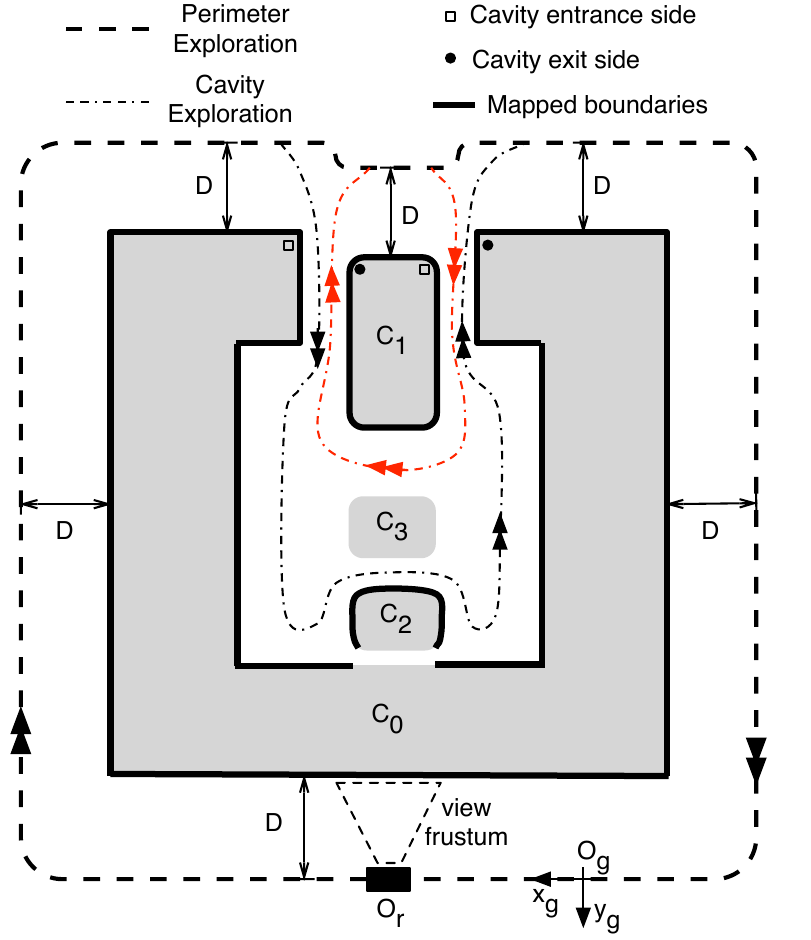}
\caption{Overview of the two phases of the mapping strategy. The
gray area represents the slice $\mathcal M$ in the plane 
$z^g=h_{\mathsf c}$ of the structure to map, with the assumption 
that any potential hanging structure above the white area 
leaves enough vertical clearance for the mobile ground robot to navigate.}
\label{fig: overview}
\end{figure}

We divide our mapping process into two phases, see Fig. \ref{fig: overview}.
The first is the Perimeter Exploration (PE) phase, during which 
the robot moves with the structure on its right to determine its boundaries.
The robot continuously moves towards previously unseen regions 
of the structure, with the exploration directed towards finding 
the limits of the structure and closing a first loop around it 
relatively quickly rather than trying to map all its details.
The PE phase ends when our algorithm detects that the robot has returned 
to the neighborhood of its starting point $\fr O_{\fr g}$
and the vSLAM module detects a global loop closure. 
After completing the PE phase, the system determines 
the locations of potential missing parts in the constructed 3D model. 
Next, in the Cavity Exploration (CE) phase, the system explores 
these missing parts in the model. 
The following subsections explain each step of our process in detail.

\section{Perimeter Exploration}
\label{section: PE}

In this section, we present a method to 
autonomously determine the boundaries of an unknown structure. 
From Assumptions \ref{assumption: annulus} and \ref{assumption: level ground}, 
$z^{\fr g}=0$ and $z^{\fr g}=H_{\max}$ are bounding horizontal planes 
for the model. 
The remaining problem is to determine the expansion of the structure in the 
$\vect{x}_{\fr g} \vect{y}_{\fr g}$ plane. 
To do this, the robot moves clockwise around the structure by determining online a discrete sequence of successive goals or waypoints. 
It tries to keep the optical axis of the depth sensor approximately perpendicular 
to the structure, which maximizes the depth resolution at which a given portion 
of the structure is captured, and increases the density of captured points.
It also tries to maintain the camera center $O_{\fr c}$ on a smooth path at 
a fixed distance D from the structure.

\subsection{Determination of the next goal}	\label{section: next goal}

\begin{algorithm}[t]
\caption{Algorithm for computing the next goal for the camera using the current point cloud in the camera FoR.}
\begin{algorithmic}[1]
\Function{computeNextGoal}{cloud\_full} 
\State cloud $\gets$ PCLremoveGroundPlane(cloud\_full)
\State cloud\_slice $\gets$ filterForwardSlice(cloud)	\label{algo: forward slice}
\State $\overline{p}^{\fr c} \gets$ PCLcompute3Dcentroid(cloud\_slice)	\label{algo: centroid}
\State $[\vect{v}_{1},\vect{v}_{2}, \vect{v}_{3}; \lambda_{1},\lambda_{2},\lambda_{3}] \gets$ PCA(cloud\_slice)	\label{algo: PCA plane fitting}
\State $\vect{\tilde n} \gets \vect{v}_3 - (\vect{v}_3\cdot\vect{z}_{\fr c})\vect{v}_3 $ \Comment{Projection on the $\vect{x}_{\fr c} \vect{y}_{\fr c}$ plane}
\label{algo: normal vector}
\State $\vect{n} \gets \vect{\tilde n} \; \text{sign}(\vect{\tilde n} \cdot \overrightarrow{O_{\fr c} \overline{p}^{\fr c}}); \vect{n} \gets \vect{n}/\|\vect{n}\|$
\State $\vect{r} \gets \vect{z}_{\fr c} \times \vect{n}$ 
\label{algo: direction vector}
\State $goal \gets \overline{p}^{\fr c} - D \, \vect{n} + step \, \vect{r}$	\label{algo: final goal}
\State \textbf{return} $goal,\vect{n}$
\EndFunction
\end{algorithmic}
\label{alg:gpp_computeNBV}
\end{algorithm}

The pseudo-code to determine the next 
position and orientation of the camera 
in our PE algorithm is shown 
in Algorithm \ref{alg:gpp_computeNBV}.
It takes as input the current point cloud produced by the camera in its FoR. 
For its implementation we rely on the Point Cloud Library (PCL) \cite{rusu20113d}. 

\begin{figure}[!t]
\centering
\subfloat[]{
	\includegraphics[width=1.62in]{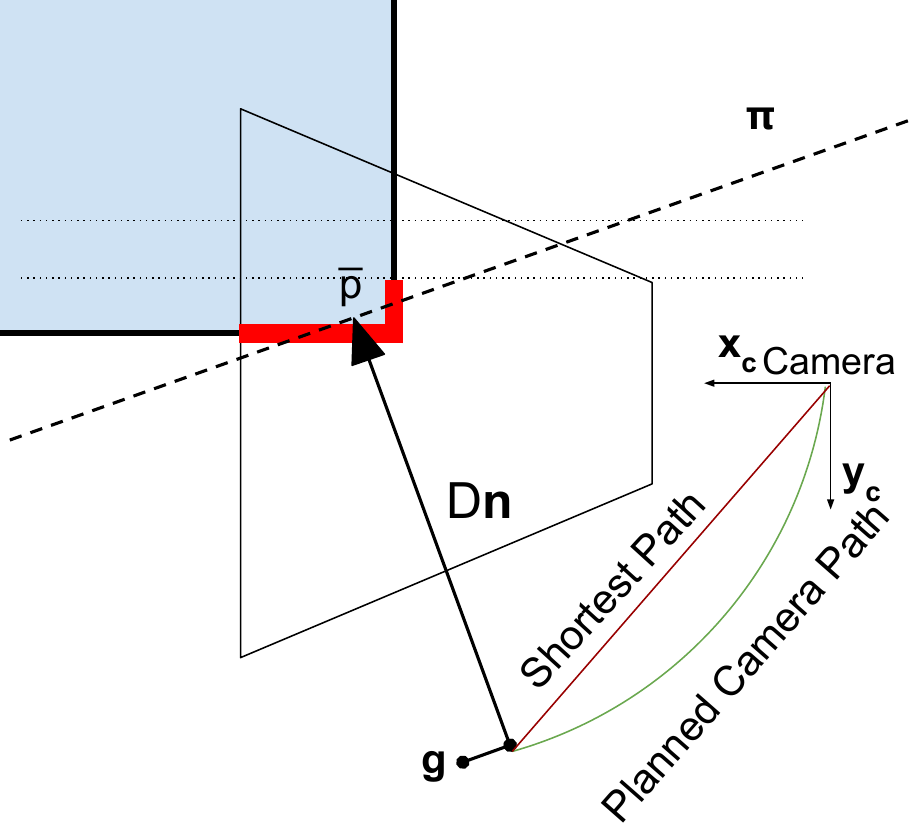}
	\label{fig:next viewpoint}
}
\hfil
\subfloat[]{
	\includegraphics[width=1.62in]{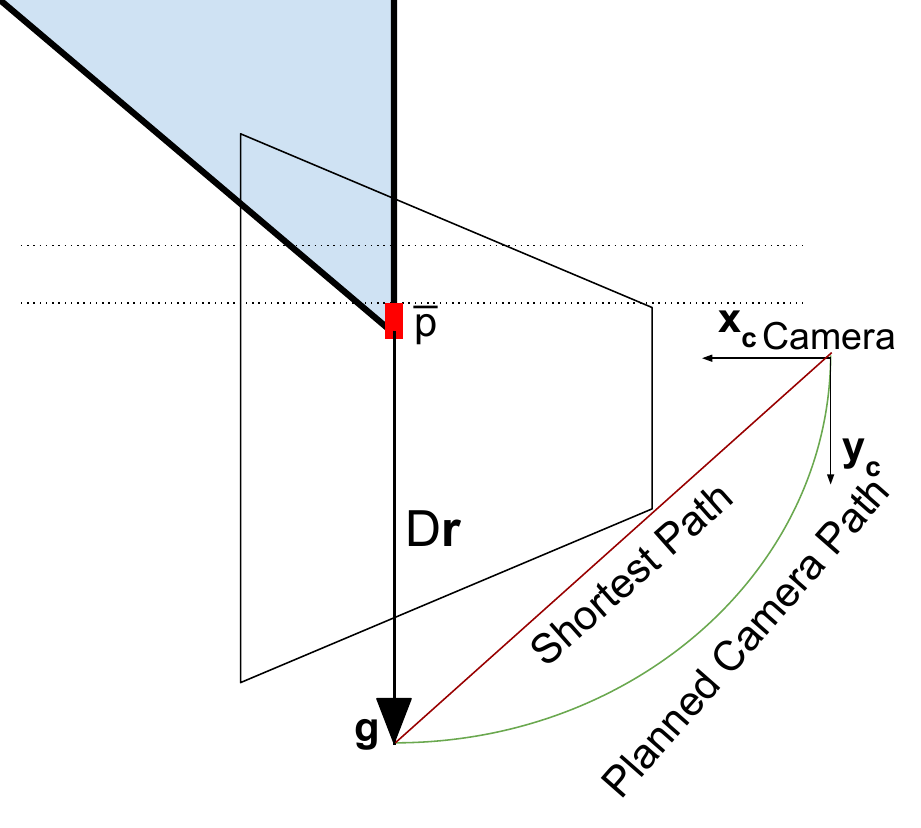}
	\label{fig:next_viewpoint_acute_angle}
}
\caption{Top-down view illustrating the computation of the next goal, for a corner
section of the structure. In (a), the forward slice $\mathcal{S}$ (highlighted in red)
contains a portion of the farther section of the corner, whereas for the acute corner
in (b), it does not and becomes very narrow.}
\label{fig:next viewpoint global}
\end{figure}

Since the next goal should depend only on the structure, we first remove 
the ground plane from the captured point cloud by removing all points below 
a certain height to obtain a point cloud $\mathcal P$.
Next, on line \ref{algo: forward slice}, we select a subset $\mathcal S$ of $\mathcal P$ 
referred to as the forward slice, which adjoins the part of the structure that must be 
explored next, see Fig. \ref{fig:next viewpoint}. 
Concretely, we choose $\mathcal S$ so that its $y^{\fr c}$-coordinates satisfy 
$y_{\max}^{\fr c} - \frac{y_{\max}^{\fr c}- y_{\min}^{\fr c}}{3} \leq y^{\fr c} \leq y_{\max}^{\fr c}$, 
where $y_{\min}^{\fr c}$ and $y_{\max}^{\fr c}$ are the minimum and maximum $y^{\fr c}$-coordinate 
values for all points in $\mathcal P$. 
On line \ref{algo: PCA plane fitting}, following \cite{mitra2004estimating}, we compute via 
Principal Component Analysis (PCA) 
the normal direction to that plane $\Pi$ which best fits $\mathcal S$.
In more details, denote $\mathcal S =\left\lbrace p_i^{\fr c} : i = 1,2,\ldots,m\right\rbrace$ 
and define the covariance matrix $\mathbf{X} = \frac{1}{m}\sum_{i=1}^{m} 
(p_i^{\fr c}-\overline{p}^{\fr c})(p_i^{\fr c}-\overline{p}^{\fr c})^{T}$, 
where $\overline{p}^{\fr c} = \frac{1}{m} \sum_{i=1}^m p_i^{\fr c}$ is the centroid 
of $\mathcal S$ computed on line \ref{algo: centroid}. 
We compute the eigenvectors $[\vect{v}_1,\vect{v}_2,\vect{v}_3]$ of $\mathbf{X}$, 
ordered here by decreasing value of the eigenvalues $\lambda_1, \lambda_2, \lambda_3$. 
The eigenvector $\vect{v}_3$ for the smallest eigenvalue corresponds 
to the normal to the plane $\Pi$.

The algorithm returns $\vect{n}$, computed from the projection of the normal vector $\vect{v}_3$
on the $\vect{x}_{\fr c} \vect{y}_{\fr c}$ plane, and taken to point 
in the direction of the vector $\overrightarrow{O_{\fr c} \overline{p}^{\fr c}}$.
This vector $\vect{n}$ defines the desired orientation of the camera.
The algorithm also returns the next goal point 
$goal = \overline{p}^{\fr c} - D \, \vect{n} + step \, \vect{r}$ for the 
center $O_{\fr c}$ of the camera, where $\vect{r} = \vect{z}_{\fr c} \times \vect{n}$ 
is computed on line \ref{algo: direction vector},
and $step = \frac{y_{\max}^{\fr c}- y_{\min}^{\fr c}}{6}$.
The term $step \, \vect{r}$, which is along the plane $\Pi$, is used to shift 
the $goal$ forward so that both sections of a corner fall in the FOV of the camera, 
as in the situation shown on Fig. \ref{fig:next viewpoint}. 
This prevents the algorithm from making slow progress around corners.
Furthermore, the interior angle of a corner could be acute, as shown in Fig.~\ref{fig:next_viewpoint_acute_angle}, and consequently the farther section 
of the corner would not be visible from the camera. Such a case can be detected 
by monitoring the width of $\mathcal{S}$ to fall below a threshold. 
In this case we modify the computation of the goal to 
be $goal = \overline{p}^{\fr c} + D\, \vect{r}$ which allows the robot 
to move around sharp corners of the structure.
Finally, the computed camera pose is transformed into the global FoR to obtain the 
next goal point $g^{\fr g}$ for the camera center $O_{\fr c}$.
We simplify the notation $g^{\fr g}$ to $g$ in the following, where we work in the
global reference frame.

\subsection{Local path planning to the next goal}	\label{section: local path planner}

\begin{figure}[!t]
\centering
\includegraphics[width=0.6\linewidth]{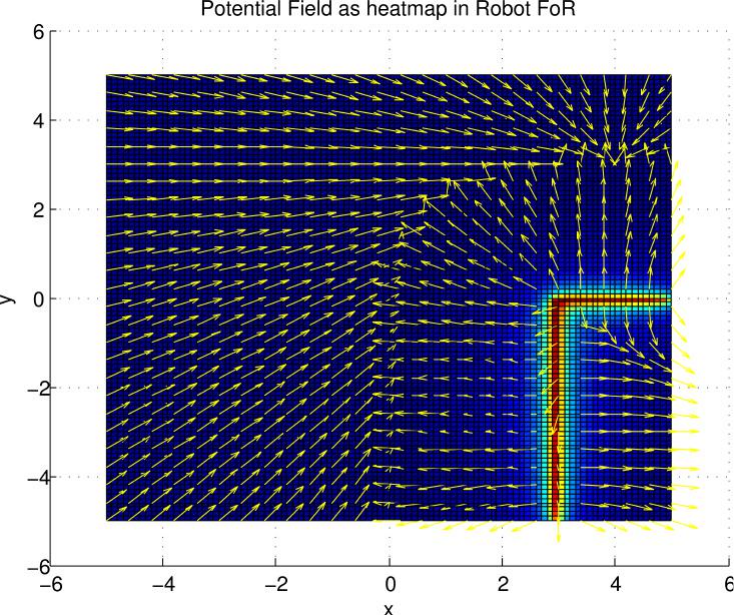}
\caption{The potential field for a goal at $(4,3)$ with $D=3$ is shown 
as a heat map and the corresponding gradient vectors are shown as a vector field.
}
\label{fig:navigation function}
\end{figure}

In order to move the camera center to $g$ while keeping it approximately
at the desired distance $D$ from the structure along the way, we use a  
local path planner based on potential fields \cite{khatib1986real, Choset_2005_5167}. 
A potential function encoding the structure as obstacles in the neighborhood 
of the camera, as well as the goal $g$, is sampled in the form of a cost map on a
local 2D grid of size $2D \times 2D$ 
centered on the camera's current position, see Fig. \ref{fig:navigation function}.
Assumption \ref{assumption: annulus} guarantees that all 
the occupied cells in this cost map denote the structure itself. 
For $k$ occupied cells centered at $\{x_j\}_{j=1}^k$, the potential function $N(x)$ is defined as
\begin{equation} 		\label{eq: N definition}
N(x) = \alpha \|x-g\|^2 + \sum_{j=1}^{k}I_j(x) d_j(x),
\end{equation}
\begin{equation*}
\text{with } d_j(x) = \frac{1}{\beta\|x-x_j\|}; 
I_j(x) = 
 	\begin{cases} 
	    1 & \text{if } \| x - x_j\| \leq D \\
		0 & \text{otherwise,}
	\end{cases}
\end{equation*}
for some scalar parameters $\alpha, \beta$.
Here $d_j$ is the repulsion from the $j^{th}$ occupied cell, 
and is limited by $I_j$ to a neighborhood of radius $D$ around the cell.
A path for the camera is obtained by following the negative gradient of $N$, i.e., 
$\dot x = -\nabla N(x)$.
Denoting $J_{x} = \left\lbrace j : I_j(x) = 1 \right\rbrace $ the occupied cells in the 
$D$-neighborhood of $x$, we have
\begin{equation}
-\nabla N(x) = 2 \alpha (g-x) + \sum_{i \in J_{x} } \frac{1}{\beta\|x-x_i\|^{3}}(x-x_i).
\label{eq:gradient}
\end{equation}

Let $\mathcal M_D = \left\lbrace x : J_{x} \neq \emptyset \right\rbrace $ denote the region 
that is at distance at most $D$ from the structure. Assuming a small value of $\beta$, the
summation term in \eqref{eq:gradient} is dominant whenever $x \in \mathcal M_D$ and pushes
the path away from the structure. However, this term vanishes as soon as $x \notin \mathcal M_D$.
Then, assuming that the camera starts at $x_0$ on the boundary $\partial \mathcal M_D$ 
of $\mathcal M_D$, it remains approximately on $\partial \mathcal M_D$ if $\overrightarrow{xg}$ 
points toward 
the interior of $\mathcal M_D$. 
It is possible that this condition is not satisfied by the point $g$ computed in the 
previous subsection, in which case we replace $g$ by $g_1$, which is obtained by 
selecting a new $goal = \overline{p}^{\fr c} - D' \, \vect{n} + step \, \vect{r}$ 
for $D'<D$ such that this condition is satisfied. The path will then slide on $\partial \mathcal M_D$
until it reaches its goal \cite{Cortes:IEEE08:nonsmoothDS}.
Finally, this path for the center of the camera is used to compute 
a corresponding path for the center of the robot, which then needs to be 
tracked using a platform specific controller.

Overall, during the PE phase the robot attempts to maintain 
a viewpoint orthogonal to the structure, even though it replans for a new goal according
to Algorithm \ref{alg:gpp_computeNBV} only at discrete times.
Note that only the computation of the next goal happens at discrete instants but the vSLAM 
module updates the model at a higher rate as per the capabilities of the hardware.

\subsection{Replanning due to the structure interferring} \label{section: obstacle ahead}


\begin{figure}[!t]
\centering
\subfloat[ ]{
	\includegraphics[width=1.5in]{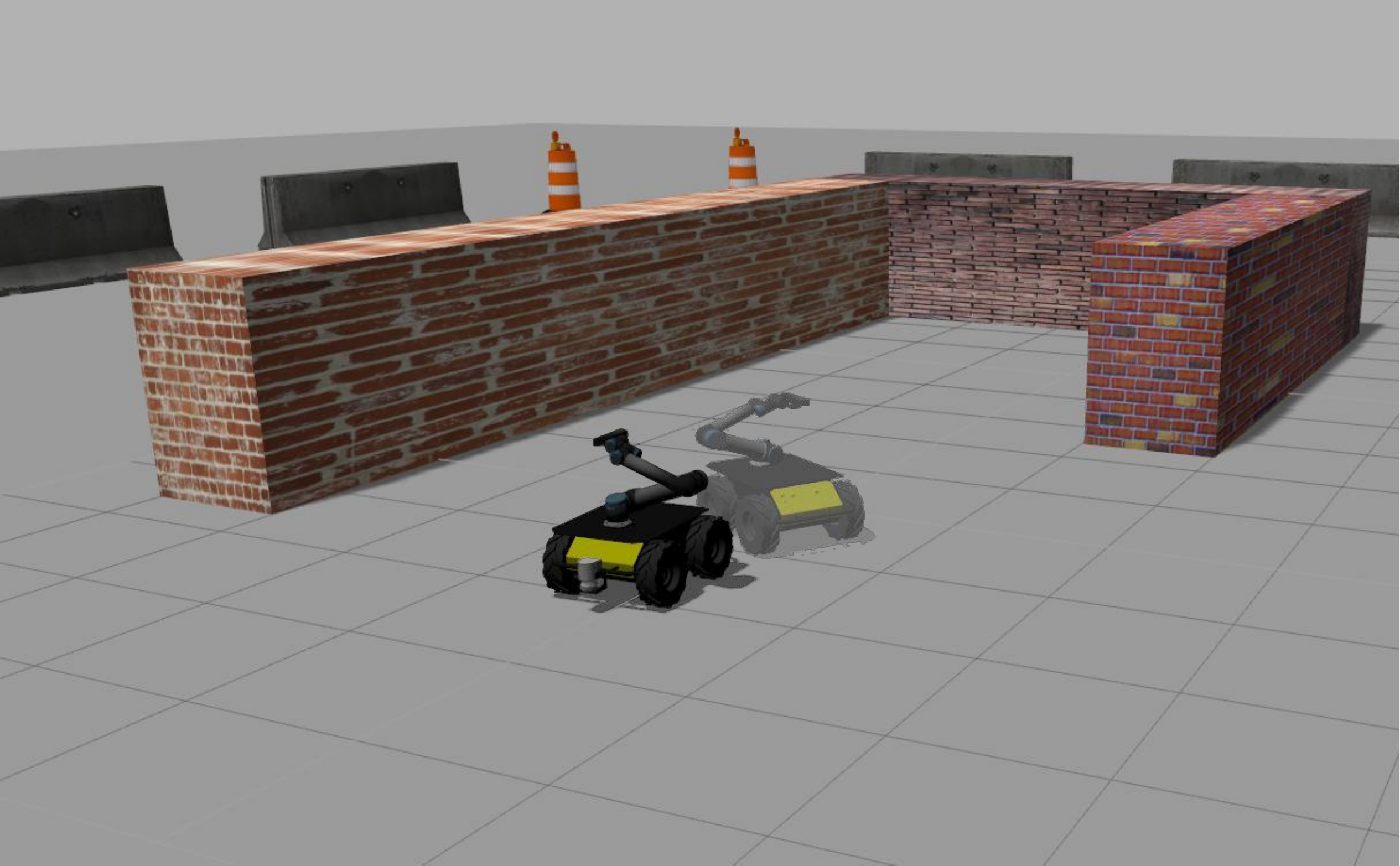}
	\label{subfig:look_ahead_gazebo}
}
\hfil
\subfloat[]{
	\includegraphics[width=1.5in]{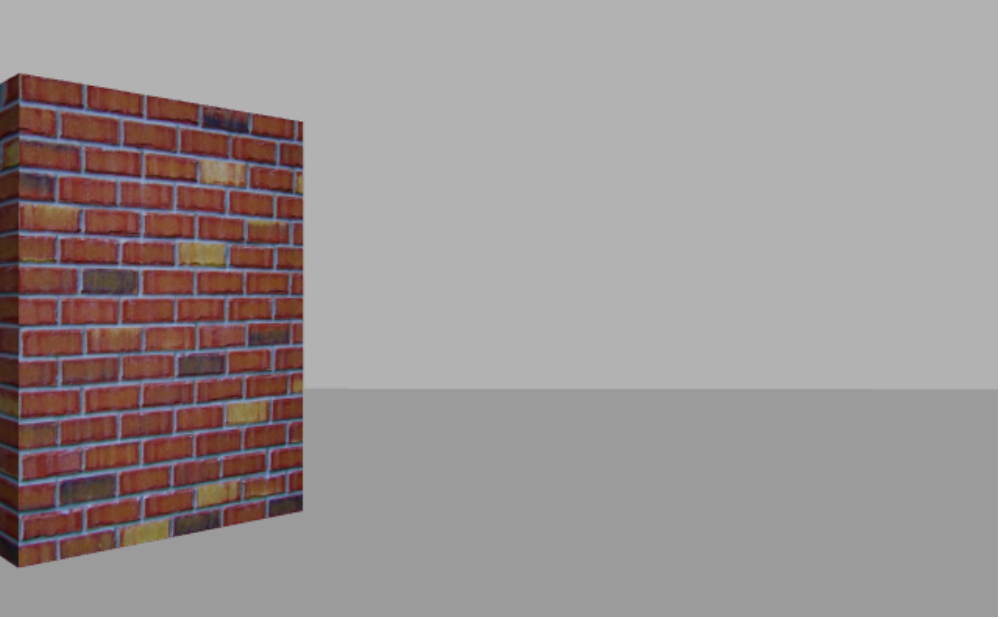}
	\label{subfig:look_ahead_camera}
}
\caption{ 
a) While the robot is following the structure, its forward facing sensors detect
an obstacle ahead (robot configuration shown in faded colors). This obstacle is 
outside the field of view of the camera, shown in b).
The position of the robot at the next waypoint along the new direction to explore, 
determined by using the arm to scan ahead, is shown in bright colors.}
\label{fig:exploration}
\end{figure}


\begin{figure}[!t]
\centering
\includegraphics[width=0.7\linewidth]{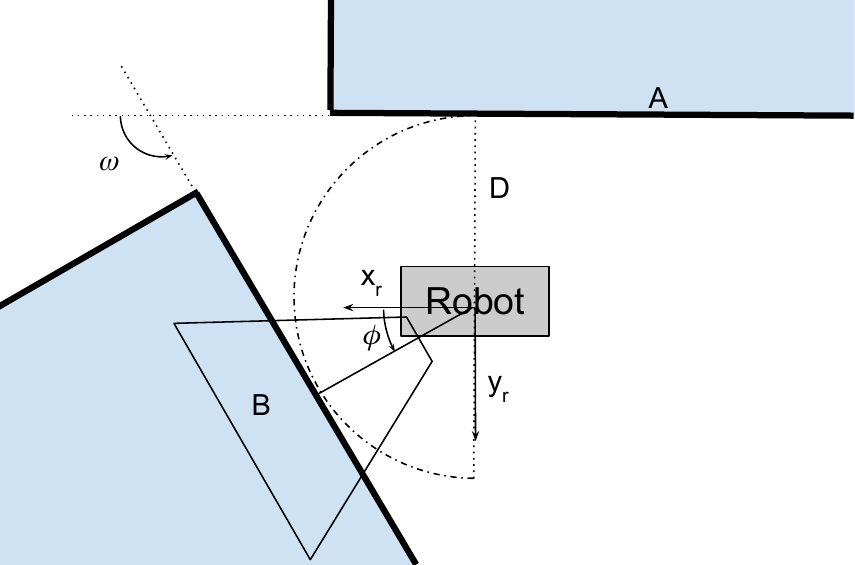}
	\caption{When the robot is currently following section A of the structure, a later section B of the structure 
could interfere with the planned path. 
The angle $\omega$ made by section B with respect to section A satisfies $\omega \in [0,\pi)$. 
Also, the two sections could be connected to form a non-convex corner.
}
\label{fig:concave_surfaces}
\end{figure}

Assumption \ref{assumption: annulus}
guarantees that the robot can move sufficiently freely around the structure, but this does not
prevent the structure itself from interfering with the path planned above.
Consider the situation shown in Fig. \ref{subfig:look_ahead_gazebo}. The wall ahead of the robot does 
not fall into the FOV of the camera due to the limited horizontal angle of view, yet the robot should not approach 
this wall closer than a distance $D$.
Hence, if the robot detects obstacles in its forward $D$-neighborhood, it is stopped at
its current position and the yaw motion of the camera is used to scan ahead and face the 
new section of the structure. 
More precisely, as illustrated in Fig. \ref{fig:concave_surfaces}, we use the costmap from the previous subsection 
to turn the camera to face along the direction from the robot center $O_{\fr r}$ to the first occupied cell in 
the $D$-neighborhood of the robot. The next goal is then recomputed using the newly 
captured point cloud.

\subsection{End of the PE phase}	\label{section: PE termination}

The end of the PE phase corresponds to the robot closing a loop around the structure. 
Therefore, we require that the vSLAM module detects a global loop closure based on 
the captured images, i.e., recognizes that the robot has returned to the vicinity 
of a known point.
The robot continues traveling on the PE path until this condition is met. 
Detecting a global loop closure is not necessarily straightforward because of localization 
errors, notably the drift accumulating in dead-reckoning systems such as the visual 
odometry function of the vSLAM module, or the wheel odometry system. 
However, it is typically possible to place a unique object or mark on or near
the structure in the initial FOV, which helps prevent incorrect loop closures.
If available, absolute positioning sensors such as a GPS receiver in the case 
of outdoor operations can also indirectly help improve the loop closure detection
by limiting the localization drift.
One can also use the measurements of a compass to detect when the robot is
traveling along an edge of the structure that has the same orientation as the starting
edge, and focus the search for a loop closure along these edges.


\section{Completing the Model: Cavity Exploration}
\label{section: CE}

There are two possible types of flaws in the model obtained at the end 
of the PE phase. Type I flaws correspond to holes that are present 
in the already explored regions. As noted in Section II, these holes 
could be due to limitations of the sensor or local occlusions caused 
by small irregularities in the structure itself, and should be filled 
using a platform with a more appropriate reachable space, hence we 
do not consider them further. 
Type II flaws, called cavities in the following, correspond to regions 
that were skipped during the PE phase, due to the situation depicted 
on Fig. \ref{fig:exploration} in particular. 
These cavities will be filled during the CE phase, where the robot 
is allowed to move closer to the structure, although this means that 
the model will not necessarily be reconstructed up to a height 
$H_{\max}$ in some places.

\subsection{Cavity Entrances}	\label{section: cavity entrance determination}

In this subsection we describe an algorithm to determine the locations 
of the entrances of the cavities in the model, which will be subsequently 
used by the CE strategy.
We use a voxel based 3D occupancy grid constructed from the global point 
cloud, and maintained in a hierarchical tree data structure by 
the OctoMap \cite{hornung13auro} library.
Internally, this library performs ray casting operations, labelling the  
occupancy measurement of each voxel along the line segment from the camera 
position to each point in the point cloud as \emph{free} and the point itself 
as \emph{occupied}.
For this, we require the vSLAM module to provide the sequence of point clouds
and associated estimated camera positions used in assembling the current model.
All voxels in the occupancy grid that are not labeled free or occupied 
are called \emph{unknown}.
Using the constructed OctoMap, we compute a set of \emph{frontier} voxels,
whose definition is adapted from \cite{Yamauchi:1997:FAA:523996.793157}.
\begin{defi}\label{defi: frontier voxel}
A frontier voxel is a free voxel with at least one neighboring unknown voxel. 
\end{defi}

\begin{figure}[!t]
\centering
\subfloat[With frontier voxels]{
	\includegraphics[width=0.3\linewidth]{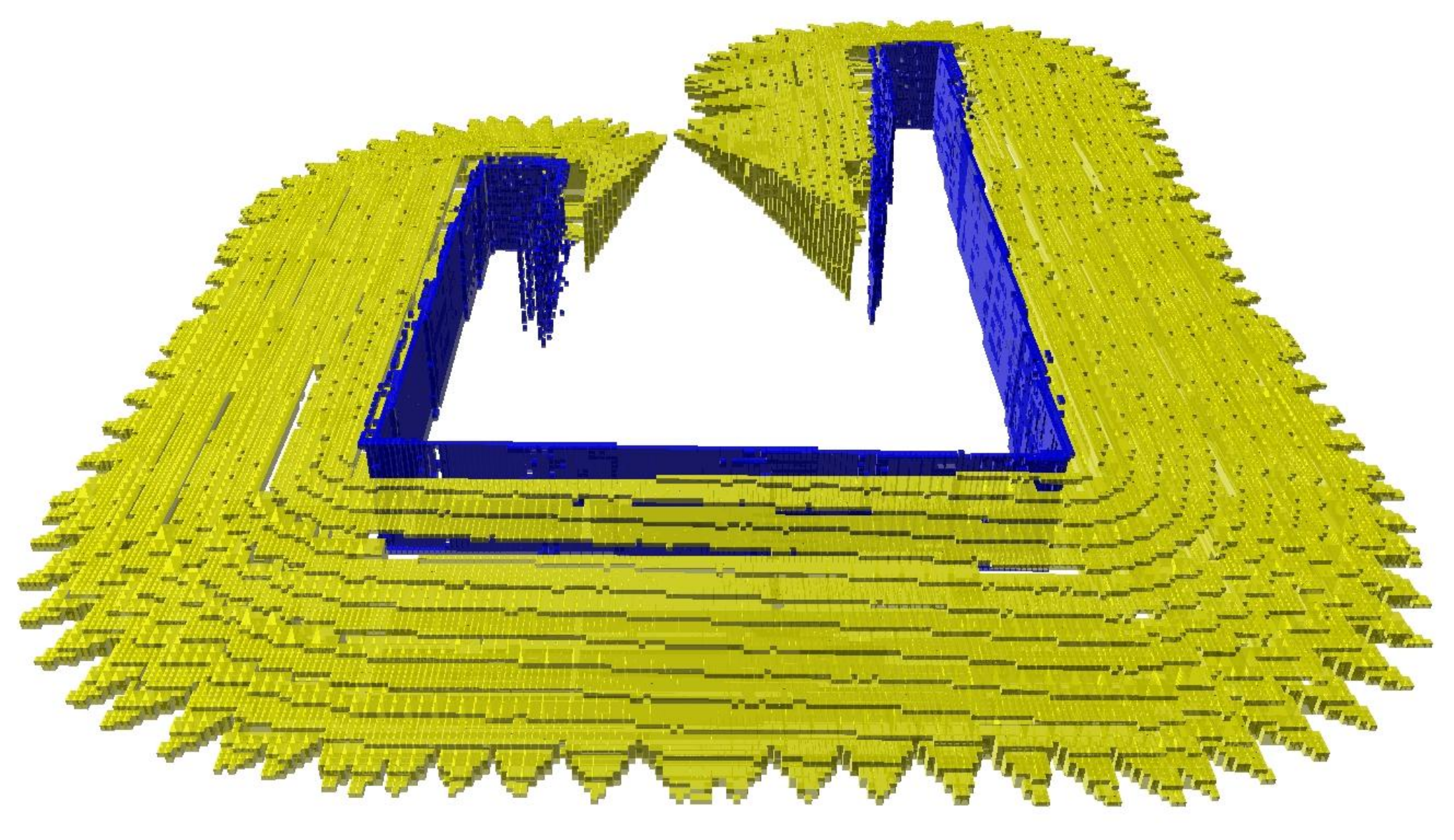}
	\label{subfig:frontier candidates}
}
\hfil
\subfloat[Cavity entrance voxels shown in red]{
    \includegraphics[width=0.6\linewidth, trim=0 10cm 0 0, clip=true]{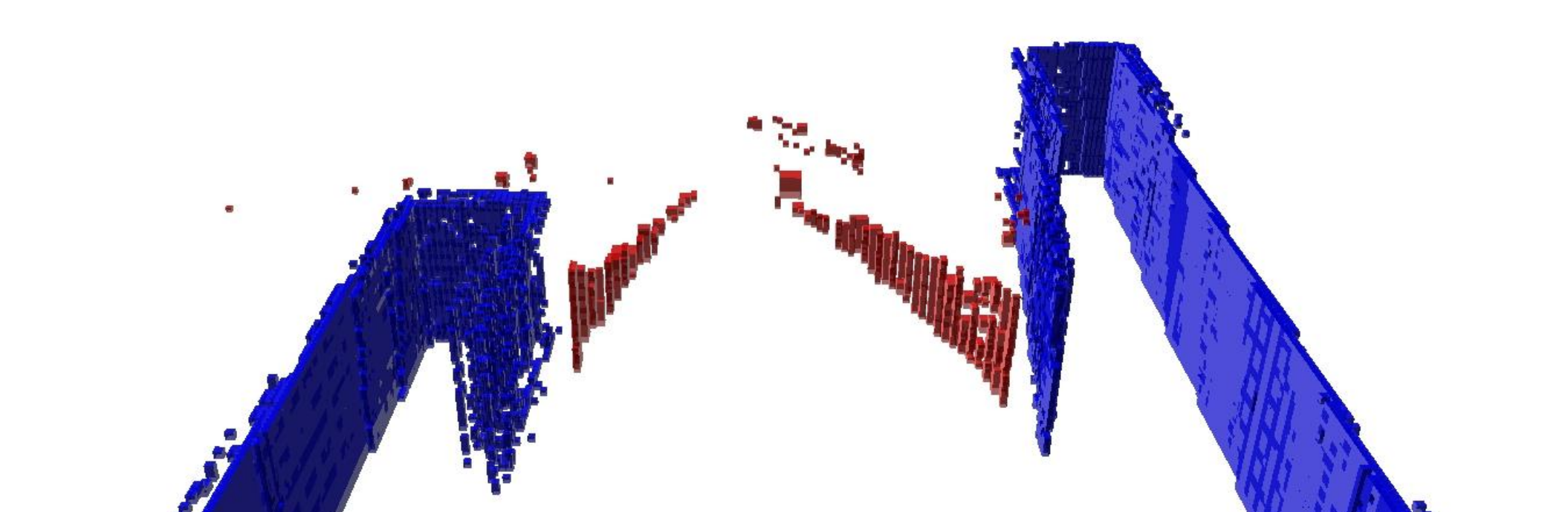}
	\label{subfig:frontier}
}
\caption{ 
a) The constructed OctoMap with occupied voxels shown in blue and frontier voxels 
shown in yellow. These yellow voxels form the boundary of the explored region, 
but most of them lie along the top and bottom faces of the view frustums. 
b) The cavity entrance voxels are shown in red.
}
\label{fig:octomap frontier}
\end{figure}

\begin{figure}[!t]
\centering
\subfloat[Leaf size: $0.05m$, depth: 16]{
	\includegraphics[width=1.5in]{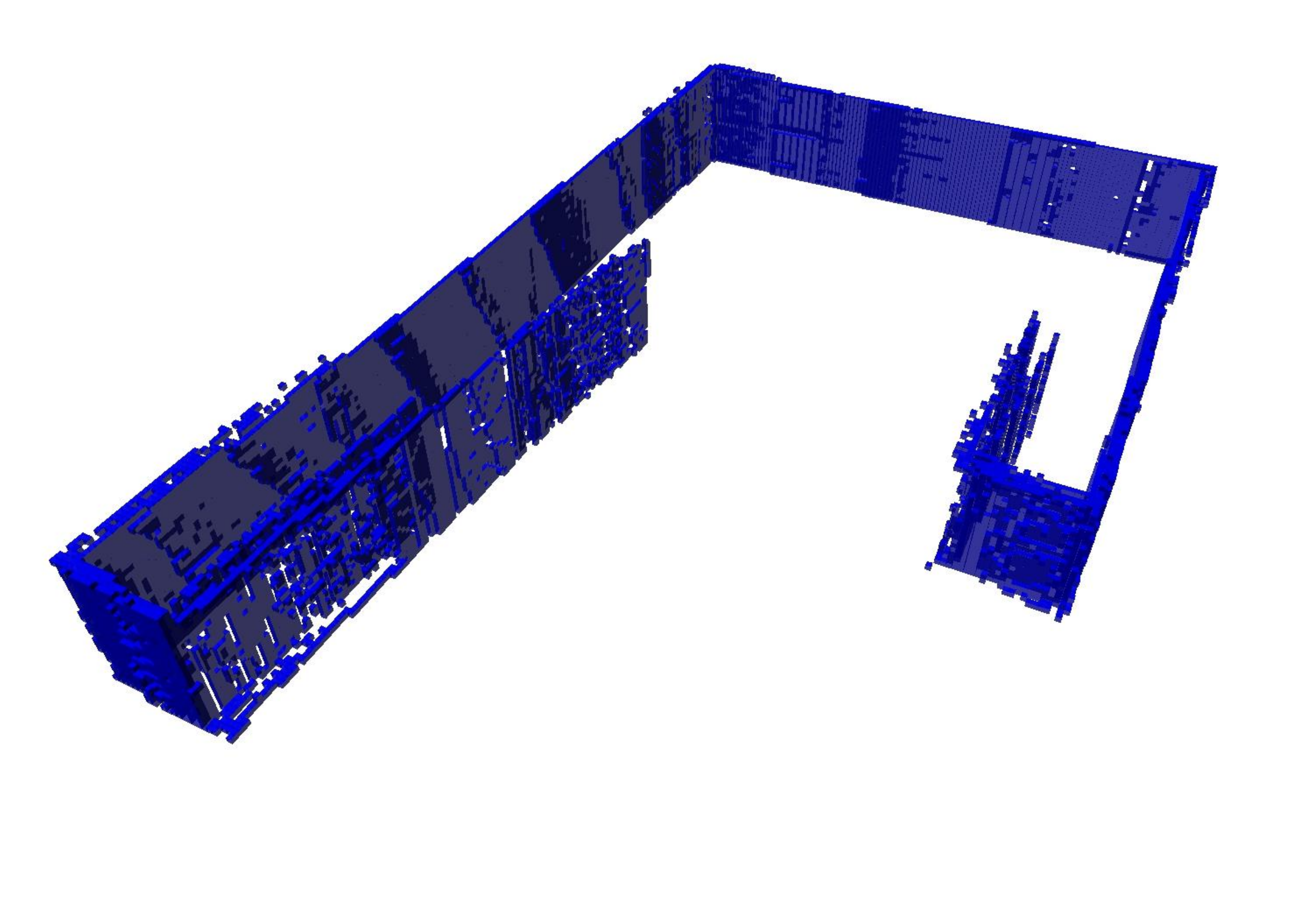}
	\label{subfig:octomap_16}
}
\hfil
\subfloat[Leaf size: $0.4m$, depth: 13]{
	\includegraphics[width=1.5in]{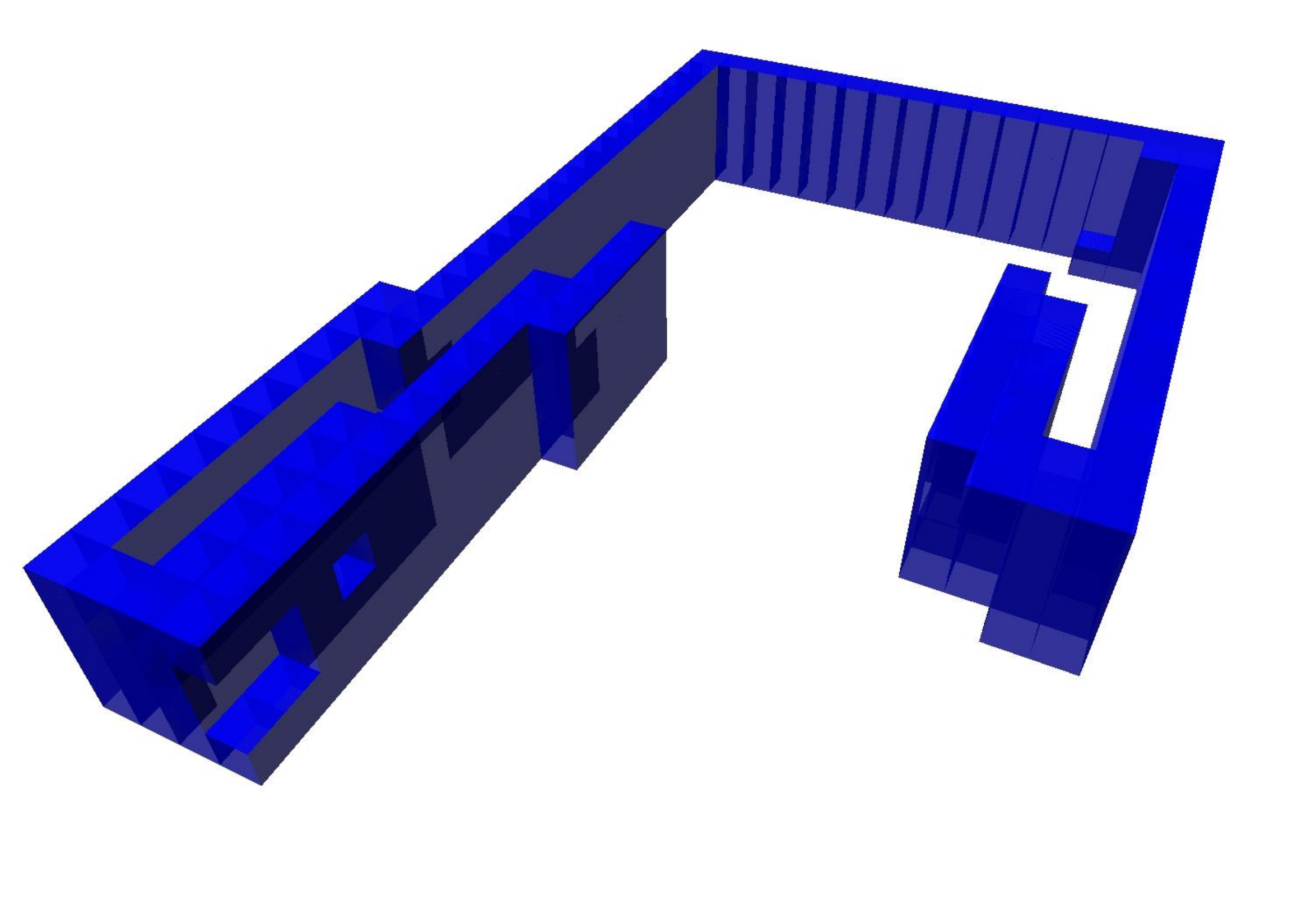}
	\label{subfig:octomap_13}
}
\caption{
OctoMap queried at depth level $16$ and $13$ respectively. In this paper, the
value of $D$ is $3\text{m}$ and the threshold $d_0$ is chosen as $0.4\text{m}$.
}
\label{fig:octomap_diff_res}
\end{figure}

Recall that the camera is constrained to move in a horizontal plane 
during the PE phase. Consequently, many frontier voxels lie along the 
top and bottom faces of the view frustums, see Fig. \ref{subfig:frontier candidates},
but do not correspond to cavities to explore. We can ignore them by 
only considering frontier voxels for which the normal vector $\mathbf{n}$, 
computed using the nearby frontier voxels\cite{mitra2004estimating}, makes
a sufficiently small angle with the horizontal plane. In other words, we keep
only the frontier voxels for which the $z$-coordinate of the normal $\vect{n}$ 
satisfies $|n^g_z| < \alpha$, for some chosen threshold $\alpha$.
Next, Type I flaws can result in frontier voxels, which we also want to exclude
from consideration. 
Therefore, we require that the distance to the closest occupied voxel 
should be greater than some threshold $d_0$, which can be chosen as  
a small fraction of the distance maintained from the structure, say $0.1\,D$.
As the number of voxels in a typical structure is very large, we do not
perform this thresholding exactly but instead we use 
an estimate for the distance to the structure obtained from OctoMap. The 
hierarchical structure of OctoMap allows efficient multi-resolution queries, 
see Fig. \ref{fig:octomap_diff_res}, and thus we keep as cavity entrance
voxels only those that are marked free at a resolution of approximately $d_0$.

Finally, we call \emph{cavity entrance voxels} the frontier voxels that 
satisfy the two preceding conditions, see Fig. \ref{subfig:frontier}. 
The cavity entrance voxels are clustered using an Euclidean clustering 
algorithm from PCL \cite{rusu20113d} and each cluster is referred to as a \emph{cavity entrance}.  
Moreover, there could be some sparsely located cavity entrance voxels, 
which are removed by setting a minimum size for the cavity entrance.

\subsection{Cavity Exploration}	\label{section: CE path planner}

\begin{figure}[!thb]
	\centering
	\includegraphics[width=\linewidth]{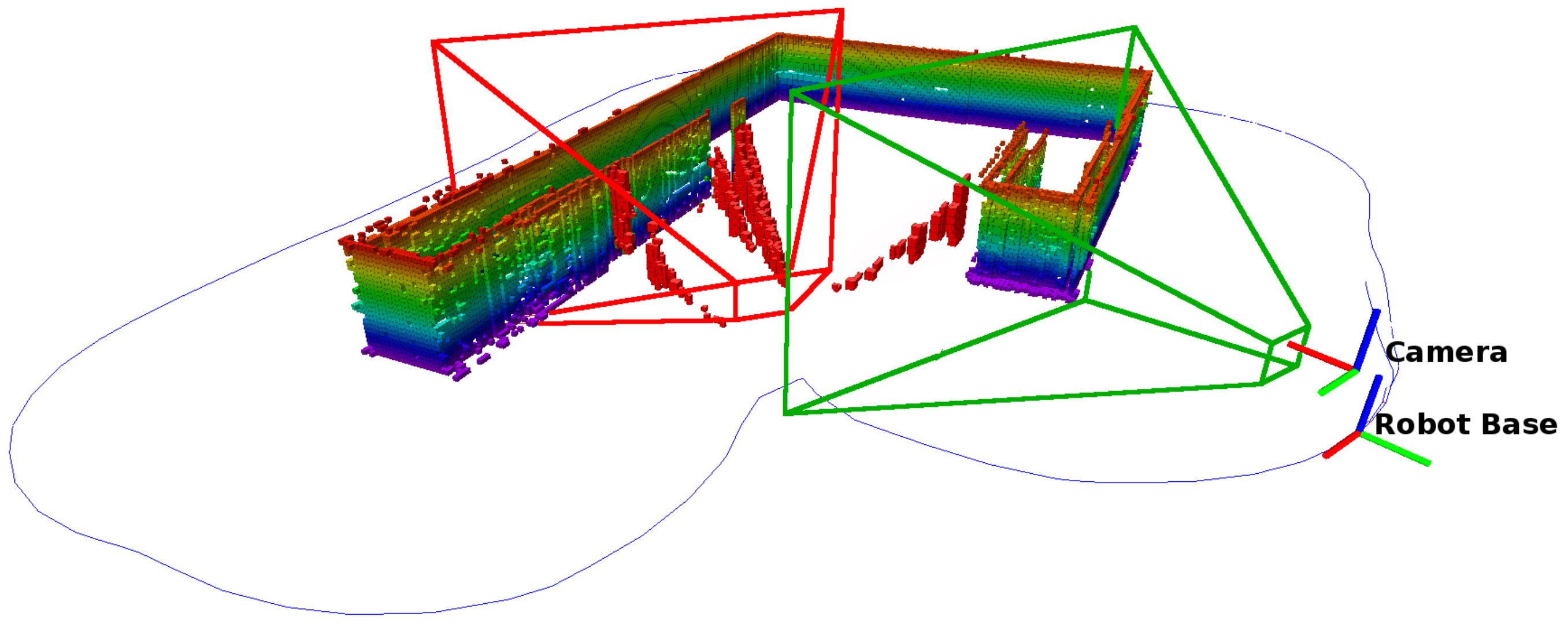}
\caption{Robot at a starting viewpoint for exploring a cavity. 
All the cavity entrance voxels are also shown. 
We also show on the left an ending viewpoint, where the vSLAM system detects
that it is back in a region already explored during the PE phase.
}
\label{fig:cavity exploration}
\end{figure}

Once the cavity entrances have been determined, we can start the CE phase. 
We explore each detected cavity using a motion analogous to 
the PE phase, wherein we maintain the structure to the right 
at a distance $\Delta \in [\delta,D]$ that is determined online based on 
the available clearance in the cavity
and $\delta$ is the minimum required distance for the mapping module. 
For this, we require a starting viewpoint for each cavity entrance 
and an algorithm to compute $\Delta$.
The starting viewpoint is chosen from the set of camera poses returned by 
the vSLAM module during the PE phase
and such that the centroid of the cavity entrance lies within the view frustum.
Additionally, the centroid should not be occluded by the structure from the
camera position.
From these camera poses, the one with the earliest timestamp 
is chosen as starting viewpoint, see Fig. \ref{fig:cavity exploration}. 

The timestamps of the starting viewpoints of the cavity entrances are 
used to sort them in increasing order and each
of the cavities is explored in sequence. 
A typical cavity has at least two cavity entrances bordering it, 
as shown in Fig.~\ref{subfig:frontier} and it is possible to have 
more cavity entrances in some cases. 
During the CE phase, if the centroid of a cavity entrance falls within 
the view frustum of the current camera position and is not occluded 
by the structure, we remove that cavity entrance from our list.

Exploring confined regions during the CE phase requires certain modifications to the
PE policy. Recall that the system skipped the cavities during the PE phase as the
robot came closer than a distance $D$ from the structure. Therefore during the CE phase, 
only the region directly ahead of the robot and within a distance $\Delta$ is checked
for interference of the computed path with the structure. Moreover, our potential field-based 
local path planner now returns paths that maintain a distance $\Delta$ from the structure.
For this, using the notation of Sections \ref{section: next goal} 
and \ref{section: local path planner}, we modify $goal$ as 
$goal \gets \overline{p}^{\fr c} - \Delta \, \vect{n} + step \, \vect{r}$ and 
transform to the global FoR to obtain the new point $g$. 
The distance $\Delta$ is chosen by starting from the minimum value $\delta$ and 
increasing it until we reach a local minimum of $N_\Delta(g)$ along $\vect{n}$, 
where the definition of $N_\Delta$ is adapted from \eqref{eq: N definition} 
with $\Delta$ replacing $D$. Similarly, when an acute angled corner is encountered 
during the CE phase, we modify \emph{goal} as 
$goal \gets \overline{p}^{\fr c} + \Delta \, \vect{r}$.

When the robot exits a cavity, the point clouds captured by the camera correspond
to parts of the structure that are already present in the model from the PE phase. 
Consequently, the system can detect that it has finished exploring the current cavity 
by monitoring the loop closures obtained by the vSLAM module. 
The robot can then choose the next region to explore from its current list 
of remaining cavity entrances, and can travel there by following again the PE path.
Alternatively, the number of changes in the occupancy measurements of the OctoMap
could be used to detect the end of the cavity, as point clouds captured after
exiting the cavity ideally would not add new information to the OctoMap. 
But this solution tends to be less robust because localization errors and 
sensor noise can induce a large number of changes even when the camera 
is viewing a region that is already present in the model.


\section{Coverage Analysis}	\label{section: analysis}

In this section we provide some analysis of the coverage completeness 
of the PE and CE strategies. 
To simplify the discussion, we focus on the case of 
simple structures consisting of vertical walls, potentially 
supporting hanging structures under which the mobile robot is
able to pass. We then analyze the boundary coverage in 2D 
for the slice $\mathcal M$ of the structure on 
the plane $z^{\fr g} = h_{\mathsf c}$, see Fig. \ref{fig: overview}.

First, we analyze the PE phase. We assume that the
path planner is able to keep the robot at distance $D$ from $\mathcal M$, in 
other words, the robot's path remains on the boundary $\partial \mathcal M_D$ 
defined in Section \ref{section: local path planner}, keeping the structure on its
right. Note that $\mathcal M_D$ is the Minkowski sum $\mathcal M \oplus \mathcal B_D$ 
of $\mathcal M$ and a closed disk of radius $D$.

\begin{lem}	\label{lem: no self-intersection}
The path followed by the robot during PE phase cannot self-intersect, except
at the initial point $O_{\mathsf g}$.
\end{lem}

\begin{figure}[!htb]
	\centering
	\includegraphics[width=0.7\linewidth]{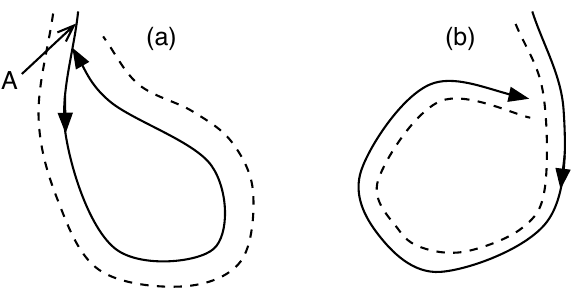}
	\caption{Impossibility of self-intersection during PE. 
    The dashed curve represents the boundary of the structure, the solid curve the path of the robot.}
\label{fig:self_intersect}
\end{figure}

\begin{IEEEproof}
During the PE phase, the robot keeps the structure at distance $D$ on its right 
as it moves forward. Fig. \ref{fig:self_intersect} then illustrates the impossibility 
for the robot's path to intersect itself during PE. Indeed, in case (a) of a 
counter-clockwise cycle, one can show that before the merging point the robot's 
obstacle detector would have seen the structure on its left at distance 
at most $D$ (point $A$ on Fig. \ref{fig:self_intersect}), and implemented 
the left turn as explained in Section \ref{section: obstacle ahead}.
In case (b) of a clockwise cycle, the tube of width $2D$ around the robot's 
trajectory would collide with the structure before closing the path, which again 
would have induced a left turn. In both cases we have a contradiction.
\end{IEEEproof}

Recall that a simple closed curve (SCC) is a non-self-intersecting, continuous loop.
We then have
\begin{cor}	\label{cor: PE path}
Suppose $\partial \mathcal M_D$ consists of a finite set of disjoint SCCs.
Then, during the PE phase, the robot travels on the SCC of $\partial \mathcal M_D$ on which 
it initially started, 
in the direction that keeps $\mathcal M_D$ on its right.
Moreover, assuming the loop closure detection does not incorrectly terminate the 
PE phase too early, the robot reaches back its starting point $O_{\mathsf g}$ on this curve.
\end{cor}

\begin{IEEEproof}
The robot progresses along $\partial \mathcal M_D$, and its path cannot
self-intersect by Lemma \ref{lem: no self-intersection}, so it must eventually 
reach back its starting point since the length of $\partial \mathcal M_D$ is finite.
It cannot switch to another SCC than the one on which it started, since it
would violate the assumption that the planner maintains a distance $D$ with
$\mathcal M$.
\end{IEEEproof}

Corollary \ref{cor: PE path} characterizes the part of the boundary of $\mathcal M_D$ 
that the PE strategy covers, assuming the path planner and PE termination algorithm work
correctly. The robot ideally travels on an SCC that is part of $\partial \mathcal M_D$, 
which we call the \emph{PE curve} in the following. We orient this curve in the direction
of travel of the robot, with $\mathcal M_D$ on the right.

\begin{figure}[!htb]
	\centering
	\includegraphics[width=0.7\linewidth]{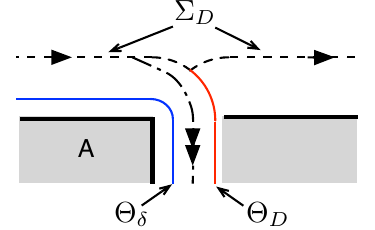}
	\caption{Illustration of the notation used in Proposition \ref{prop: PE+CE analysis}. 
    Here $\mathcal C_\delta = A \oplus \mathcal B_\delta$.}
    \label{fig: CE curves}
\end{figure}

Let us now turn to the analysis of the CE phase. 
Let $\mathcal M_\delta = \mathcal M \oplus \mathcal B_\delta$ be the Minkowski sum
of $\mathcal M$ and the closed disk of radius $\delta$, where $\delta$ is the minimum
horizontal clearance defined in Section \ref{section: CE path planner}. 
We work under the mild assumption that both $\partial \mathcal M_D$ and 
$\partial \mathcal M_\delta$ consist of a finite set of disjoint SCCs, 
although $\partial \mathcal M_D$ can have a strictly smaller number of such
curves in general. The notation of the following proposition is illustrated on Fig. \ref{fig: CE curves}.

\begin{prop}	\label{prop: PE+CE analysis}
Let $\Sigma_D$ be the oriented PE curve, and $\mathcal C_\delta$ be a connected component 
of $\mathcal M_\delta$ in the region on the right of $\Sigma_D$.
Let $\Theta_\delta$ be one of the SCC forming $\partial \mathcal C_\delta$, 
orient $\Theta_\delta$ such that $C_\delta$ is on its right,
and let $\Theta_D$ be the (possibly empty) SCC forming the boundary of   
$\Theta_\delta \oplus \mathcal B_{D-\delta}$ on the left of $\Theta_\delta$.
If $\Theta_D \cap \Sigma_D \neq \emptyset$, 
then at the end of the PE and the CE phase,  
the view frustum has covered the curve $\Theta_\delta$.
\end{prop}

Referring to Fig. \ref{fig: overview}, $\mathcal M$ has four components $C_i, i=0,\ldots,3$. 
$\mathcal M_D$ has a unique component since by adding a buffer $D$ the components merge into one.
Note however that in general, $\mathcal M_D$ does not have to be simply connected, 
nor even path connected.
The dashed line representing the PE curve is also the boundary of $\mathcal M_D$. 
Now $\mathcal M_\delta$ has three components, because $C_0$ and $C_2$ merge once
we add a buffer $\delta$. $C_1$ and $C_3$ remain disconnected in $\mathcal M_\delta$
however, which allows the robot to enter the passages separating $C_0$ and $C_1$ on the
one hand, and $C_2$ and $C_3$ on the other hand. At the end of cavity exploration, the
boundary of the components $C_1 \oplus \mathcal B_\delta$ and 
$(C_0 \cup C_2) \oplus \mathcal B_\delta$ will be mapped, but $C_3 \oplus \mathcal B_\delta$ does
not satisfy the hypothesis of Proposition \ref{prop: PE+CE analysis} (the boundary of
$C_3 \oplus \mathcal B_D$ does not share any point with the PE curve), and in this case its boundary 
indeed is not mapped. The robot cannot map the whole boundary of $C_0$ or $C_2$ individually, 
since it cannot pass between these two structures that are less than $\delta$ apart.

\begin{IEEEproof}
Note that $\Theta_D$ is a SCC that forms part or all of $\partial C_D$, where 
$C_D = C_\delta \oplus \mathcal B_{D-\delta}$, i.e., $\Theta_D$ consists of points 
that are at distance $D$ of the portion of the structure in $C_\delta$. 
As a result, all the points belonging to $\Theta_D$ must be either also on $\Sigma_D$ 
or on the right of $\Sigma_D$. A first possibility is that $\Theta_D = \Sigma_D$, 
in which case the curve $\Theta_\delta$ is covered at the end of the PE phase. 

If $\Theta_\delta$ is not covered at the end of the PE phase there is a point on $\Theta_\delta$
that lies on a cavity entrance (frontier boundary between the free and unknown region) and 
that is reachable by a path starting from $\Sigma_D$ (since $C_\delta$ is a connected component 
of $\mathcal M_\delta$, a robot could travel along $\Theta_\delta$ during the CE phase). 
Assuming this point is detected by the procedure of Section \ref{section: cavity entrance determination}, 
during the CE phase the robot will travel to this point and remove it from its list of cavities to
explore, keeping $C_\delta$ on its right along the way.
It will then continue following a path along $C_\delta$ contained in the annulus between 
$\Theta_\delta$ and $\Theta_D$, until $\Theta_\delta$ has been entirely covered 
by the view frustum. The coverage of $\Theta_\delta$ terminates since it is a SSC.
\end{IEEEproof}

In conclusion, the cavity entrances computed at the end of the PE phase act as attractors 
for the robot during the CE phase. However, the robot only covers those frontier voxels
that it can reach while still keeping the structure on its right as a guide and remaining 
at a distance between $\delta$ and $D$ away from it. 
For example, if it enters a large room after going through a cavity entrance,
it will not try to cover the area far away from the walls (hence, it does not try to
cover obstacle $C_3$ on Fig. \ref{fig: overview}). 
One could potentially attempt to cover these interior areas as well at the same time, 
e.g., by using a $2D$ coverage algorithm when we enter a wide cavity, 
but this would require in general a sufficiently precise absolute positioning 
system complementing the odometry information of the vSLAM module. This might 
not be a trivial requirement, for example because the structure itself might 
obstruct GPS reception. Instead, our algorithm is motivated by the fact that
keeping the structure in range helps maintain the accuracy of the visual 
odometry component of the vSLAM module. By trying to exit a cavity quickly 
once we enter it, the vSLAM module can also close loops more frequently as 
the robot returns to the PE curve, before accumulating too much error through 
the odometry.


\section{Simulation Results}
\label{section: validation}

We illustrate the behavior of our policies via 3D simulations for different sizes of the 
structure, camera range values and localization accuracy levels for the robot.
The implementation of our motion planning policies is 
integrated with the Robot Operating System (ROS) Navigation Stack \cite{rosnav}, 
which is supported by many mobile ground robots.
All the simulations are performed using the Gazebo simulator \cite{gazebo}. 
The vSLAM algorithm used is RTAB-Map \cite{labbe14online}.

The simulations are carried out with publicly available models of a 
Clearpath Husky A200 robot and a Kinect depth sensor whose range
can be varied \cite{gazebo_plugins}, see Fig. \ref{fig:initial conditions}. 
A UR5 robotic arm is used to carry the sensor, but only yaw motions of the 
arm are allowed, as described in Section \ref{section: statement}.
For illustration purposes, we consider artificial structures made of short wall-like 
segments. We refer to the structure used in most of the previous illustrations as the 
Small $\Gamma$ model. The Large $\Gamma$ model has the same shape as Small $\Gamma$ 
but is twice the size. We also illustrate the effectiveness of our policy
for a realistic model of a house, and compare its performance with that of the 
classic Frontier-Based Exploration (FBE) algorithm \cite{Yamauchi:1997:FAA:523996.793157}.
We have included a supplementary MP4 format video, which shows the simulation and real-world
experiments with a Husky robot following our policies for mapping the Small $\Gamma$ model 
using a Kinect sensor.

\subsection{Structure Size and Camera Range}

\begin{table}[!t]
\renewcommand{\arraystretch}{1.3}
\caption{Simulation results for different sizes of the structure and range of the camera}
\label{tbl: simulation results}
\centering
\begin{tabular}{|c|c|c|c|c|}
\hline
Model & Perimeter & Camera Range & Path Length \\
\hline
Small $\Gamma$  & $42\text{m}$ & $4.5\text{m}$ & $72.08\text{m}$  \\
Small $\Gamma$  & $42\text{m}$ & $12.0\text{m}$ & $53.79\text{m}$ \\
Large $\Gamma$  & $84\text{m}$ & $4.5\text{m}$ & $106.23\text{m}$ \\
Large \large{a} & $94\text{m}$ & $4.5\text{m}$ & $179.65\text{m}$ \\
\hline
\end{tabular}
\end{table}

\begin{figure}[t]
\centering
\subfloat[Model: Small $\Gamma$, Range: $4.5m$]{
    \includegraphics[width=1.3in]{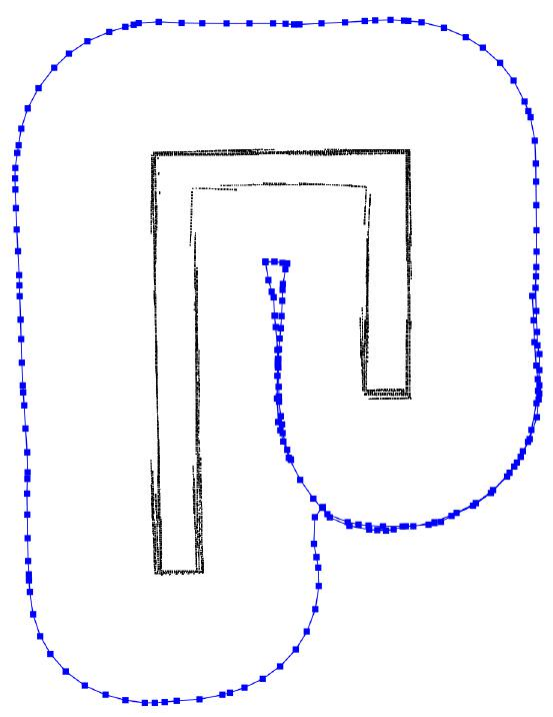}
	\label{subfig:small gamma husky 4.5}
}
\hfil
\subfloat[Model: Small $\Gamma$, Range: $12m$]{
	\includegraphics[width=1.3in]{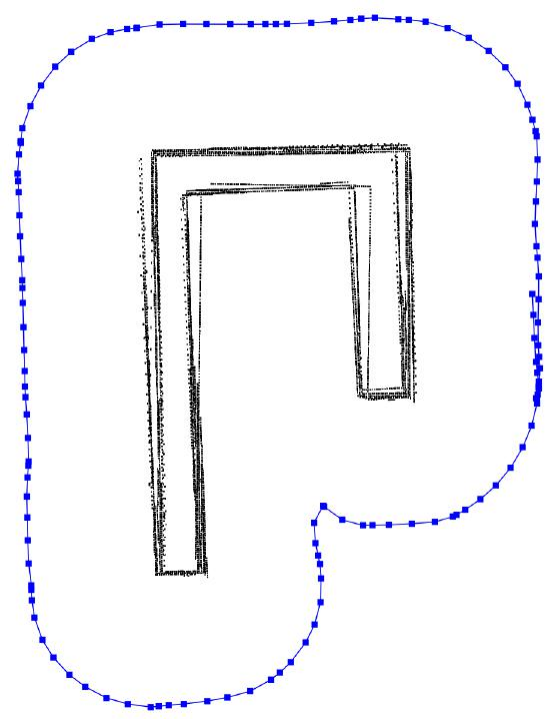}
	\label{subfig:small gamma husky 12}
}
\hfil
\subfloat[Model: Large $\Gamma$, Range: $4.5m$]{
	\includegraphics[width=1.2in]{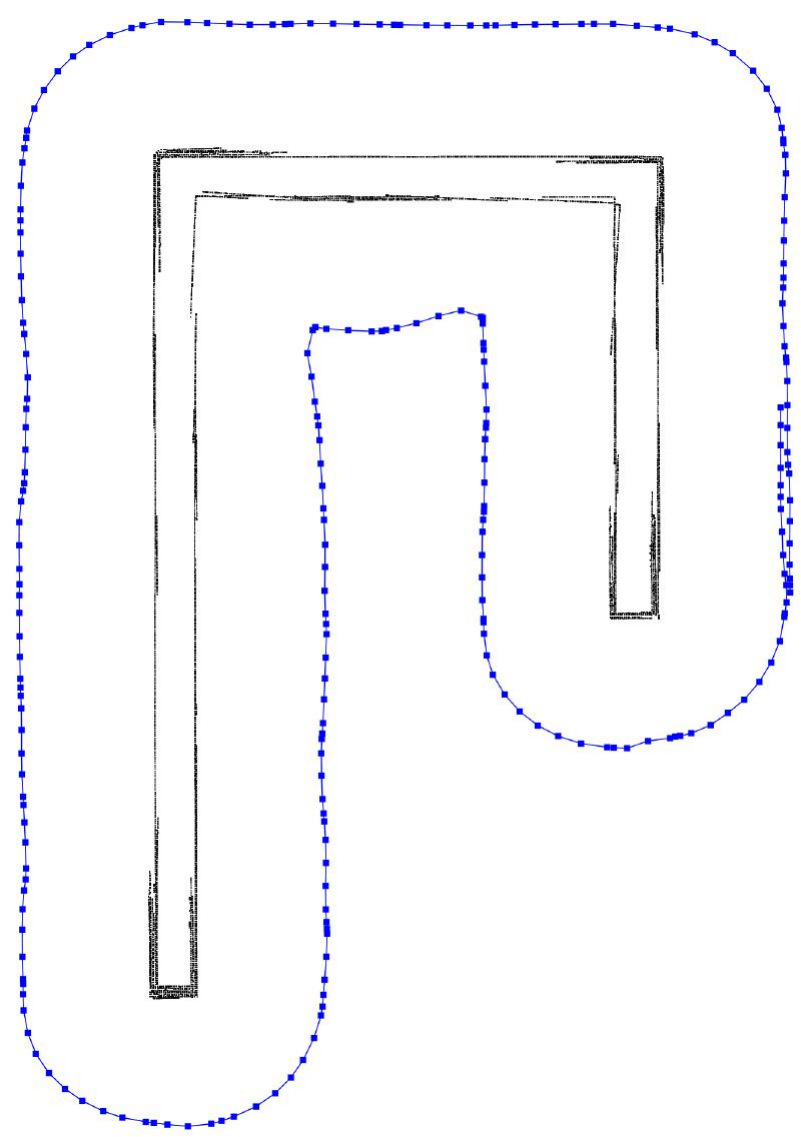}
	\label{subfig:large gamma husky 4.5}
}
\hfil
\subfloat[Model: Large {\large a}, Range: $4.5m$]{
	\includegraphics[width=1.3in]{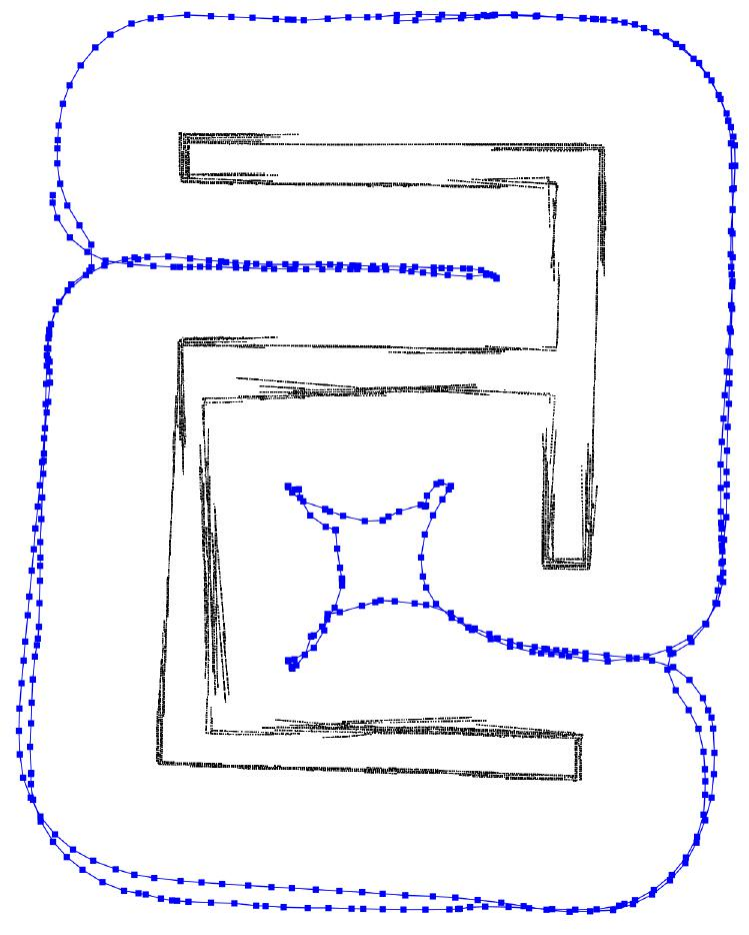}
	\label{subfig:large a husky 4.5}
}
\caption{
The projection of the reconstructed model on the $\vect{x}_{\fr g}\vect{y}_{\fr g}$ plane is shown in black and the trajectory followed by the robot based on our policies is shown in blue. 
}
\label{fig:relative size}
\end{figure}

The relative size of the structure with respect to the range of the camera affects the trajectory
determined by our algorithms. Fig.~\ref{fig:relative size} shows simulation results for $4$ scenarios.
With a camera range of $4.5\text{m}$, the Large $\Gamma$ model is completely mapped at
the end of the PE phase. For the Small $\Gamma$ model a cavity remains, which is subsequently
explored during the CE phase. 
Increasing the camera range to say $12\text{m}$ allows the Small $\Gamma$ structure to be mapped
at the end of the PE phase as well, see Fig. \ref{subfig:small gamma husky 12}. 
One can see on Fig. \ref{subfig:large a husky 4.5} that the robot following our policies is able 
to map large structures with multiple cavities of different sizes. 
Table \ref{tbl: simulation results} lists the path lengths obtained for the different test cases.

\subsection{Localization accuracy}

\begin{table}[t]
\renewcommand{\arraystretch}{1.3}
\caption{Simulation results for different levels of localization accuracy}
\label{tbl: simulation results 2}
\centering
\begin{tabular}{|c|c|c|c|c|}
\hline
$k$ & $n_k$ & $\mu$ & $\sigma$ & $\max$ \\
\hline
$0.00$ & $65520$  & $0.05\text{m}$ & $ 0.04\text{m}$ & $0.20\text{m}$ \\
$0.25$ & $72636$  & $0.08\text{m}$ & $ 0.06\text{m}$ & $0.27\text{m}$ \\
$0.50$ & $82728$  & $0.11\text{m}$ & $ 0.09\text{m}$ & $0.40\text{m}$ \\
$0.75$ & $111321$ & $0.16\text{m}$ & $ 0.13\text{m}$ & $0.94\text{m}$ \\
\hline
\end{tabular}
\end{table}

\begin{figure}[!t]
\subfloat[Reconstructed model with $k=0.75$]{
	\includegraphics[width=1.5in]{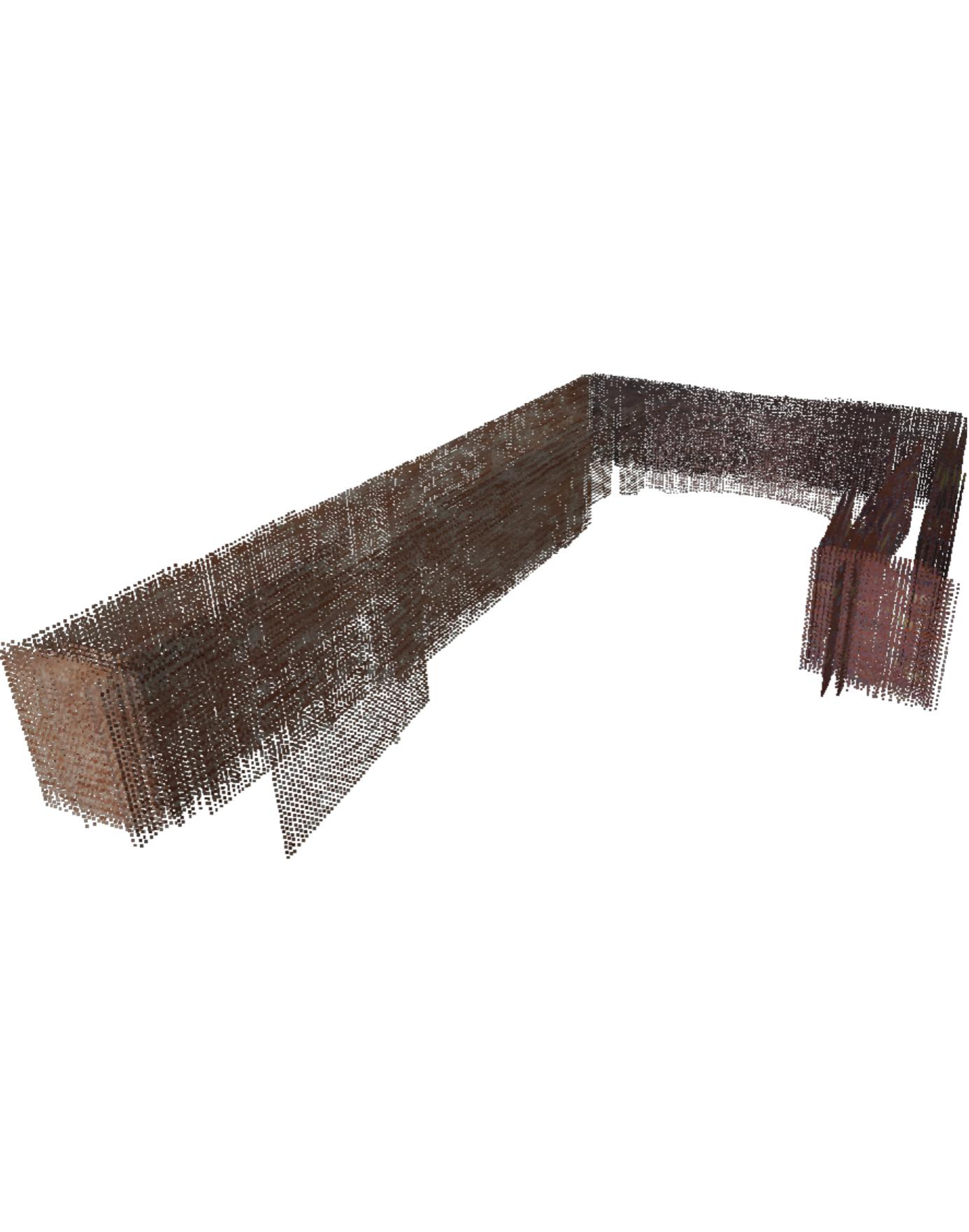}
	\label{subfig:results Husky SmallGamma 4.5 NoisyOdom Model}
}
\hfill
\subfloat[Model: Small $\Gamma$, Range: $4.5\text{m}$]{
	\includegraphics[width=1.5in]{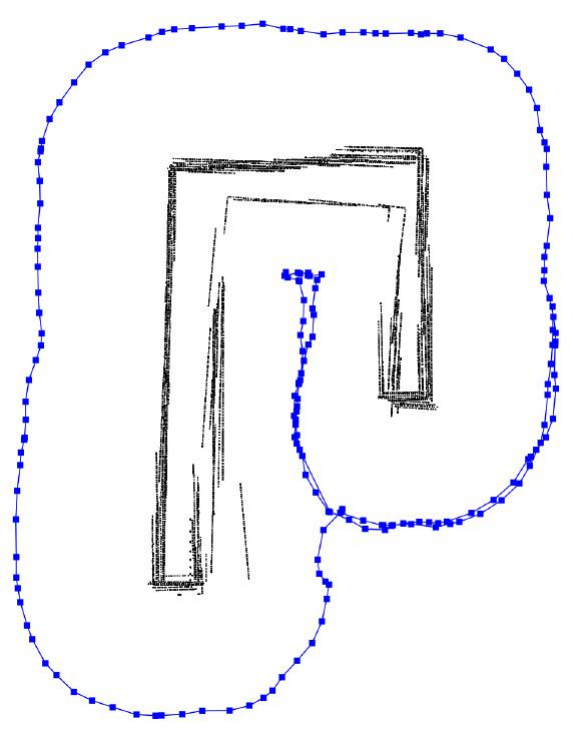}
	\label{subfig:results Husky SmallGamma 4.5 NoisyOdom}
}
\caption{a) Large errors in localization results in poor alignment, although 
all portions of the structure have been captured in the model, see 
Fig. \ref{subfig:small gamma husky 4.5 model} for comparison. b) The projection 
of the reconstructed model on the $\vect{x}_{\fr g}\vect{y}_{\fr g}$ plane, 
when compared to Fig. \ref{subfig:small gamma husky 4.5}, shows the 
distortion introduced due to the noisy wheel odometry.}
\label{fig:localization accuracy}
\end{figure}

The Husky robot combines data from an Inertial Measurement Unit (IMU), a standard GPS receiver 
and wheel odometry to achieve a relatively small localization error overall. 
In order to evaluate the impact of localization accuracy on our algorithms, we simulate 
the effect of large wheel slippage 
by introducing a zero mean additive Gaussian white noise to each of the wheel encoder measurements, with a variance equal to $k(v_x+\omega_z)/2$, where $v_x$ is the linear velocity of the robot, $\omega_z$ is its yaw rate and $k$ is a proportionality constant, also called noise level in the following.
Increasing $k$ results in a poorer alignment of the point clouds, but all portions 
of the structure, except the horizontal faces, are still captured in the reconstructed 
model, see Fig. \ref{fig:localization accuracy}. Note that our policies compute the next waypoint at 
discrete times and therefore assume that the drift in localization between waypoints 
is sufficiently small so that the robot reaches the next waypoint with the 
camera facing the structure.

We use the CloudCompare\cite{cloud_compare} software to compute the distortion 
in the reconstructed model $\mathcal{C}_k$, for a noise level $k$, with respect 
to a reference point cloud $\mathcal{C}_R$ generated using a different mobile 
platform with almost perfect localization. 
First, we register $\mathcal{C}_k$ to $\mathcal{C}_R$ using an 
Iterative Closest Point (ICP) algorithm \cite{rusinkiewicz2001efficient}.  
We then define for every point in $\mathcal{C}_k$, its error to be the distance 
to the nearest neighbor in $\mathcal{C}_R$. Table \ref{tbl: simulation results 2} 
lists the simulation results for mapping the Small $\Gamma$ model with different 
noise levels $k$, where $n_k$ is the number of points in $\mathcal{C}_k$ and 
$\mu, \sigma, \max$ are respectively the mean, standard deviation and maximum value 
of the errors of all points in $\mathcal{C}_k$. 
The table indicates that both the mean and standard deviation of the errors 
increase with the noise level.

\subsection{Comparing with Frontier-based Exploration}

The Frontier Exploration\cite{rosfrontier} package available in ROS
relies on a 2D LIDAR to build an occupancy grid that is used to compute
the frontiers. The package requires the user to define a 2D polygon 
that encloses the structure. The algorithm then explores until there
are no more frontiers inside the user-defined polygon. 
In comparison to the FBE algorithm, our algorithms
1) do not require a user defined bounding polygon;
2) maintain as much as possible a fixed distance from the structure (during the PE phase),
thereby ensuring that all portions up to a height of $H_{\max}$ are mapped;
3) consistently explore the structure while keeping it on the right, which can
be important from a user perspective to understand the behavior of the robot. 
On the other hand, the trajectory prescribed by the FBE algorithm depends on the 
size of the user-defined bounding polygon. A large bounding polygon will cause the 
robot to explore areas far away from the structure and will possibly not maintain 
a fixed direction of exploration. Our strategy produces the same path every time 
for a given structure whereas the path computed by the FBE algorithm could 
differ greatly between two trials. The robot also often gets stuck while using 
the FBE algorithm as the computed waypoints are often too close to the structure.

\begin{table}[!t]
\renewcommand{\arraystretch}{1.3}
\caption{Comparison between our policy and frontier based exploration}
\label{tbl: simulation results 3}
\centering
\begin{tabular}{|c|c|c|c|}
\cline{3-4}
\multicolumn{2}{c|}{} & Proposed Policy & FBE \\
\hline
\multirow{3}{*}{Small $\Gamma$} & Path Length            & $72.08\text{m}$  & $49.78\text{m}$  \\\cline{2-4}
								& Unique closest point set size & $6,063$          & $5,398$          \\\cline{2-4}
								& Mean Error             & $0.05\text{m}$ & $0.12\text{m}$\\\cline{2-4}
\hline
\multirow{3}{*}{House} & Path Length & $59.89\text{m}$ & $47.55\text{m}$ \\\cline{2-4}
					   & Unique closest point set size  & $9,182$ & $7,402$ \\\cline{2-4}
					   & Mean Error             & $0.05\text{m}$ & $0.12\text{m}$\\\cline{2-4}
\hline
\end{tabular}
\end{table}

\begin{figure}[t]
\subfloat[Proposed Policy]{
	\includegraphics[width=1.50in]{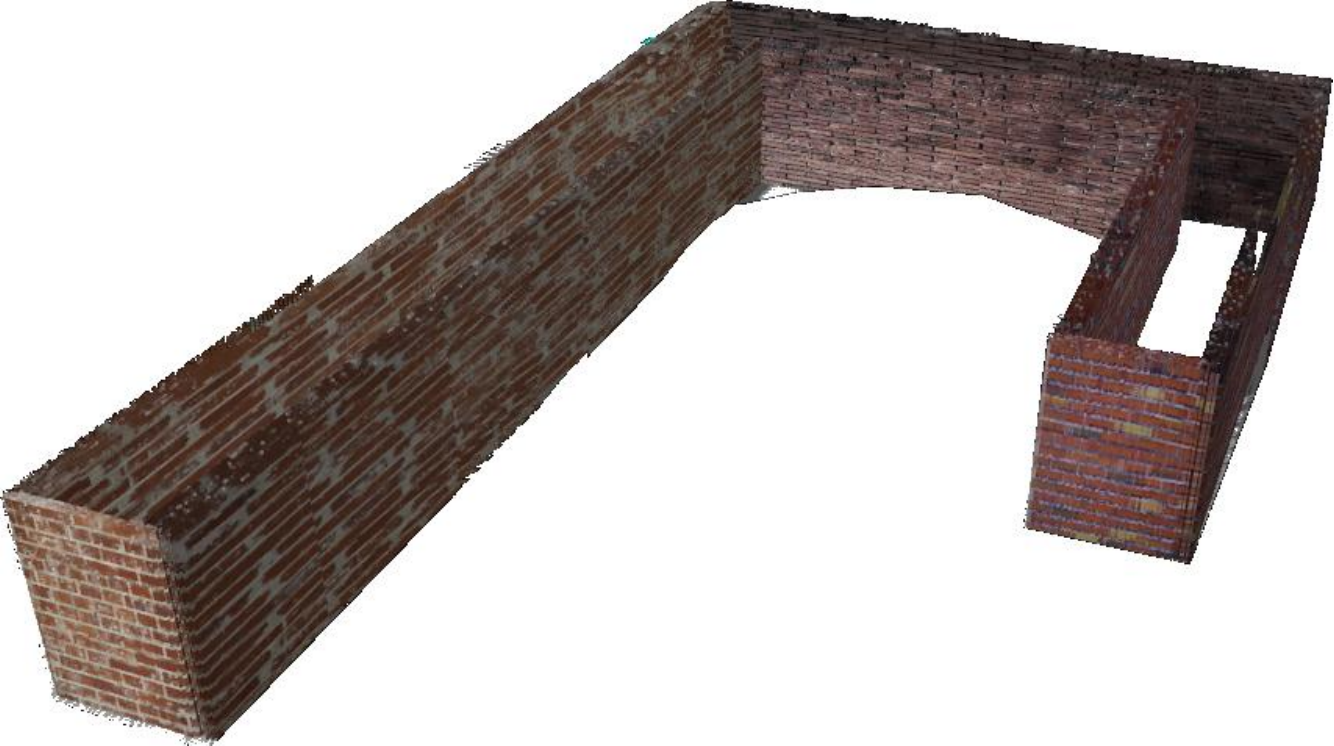}
	\label{subfig:small gamma husky 4.5 model}
}
\hfil
\subfloat[Frontier based Exploration]{
	\includegraphics[width=1.50in]{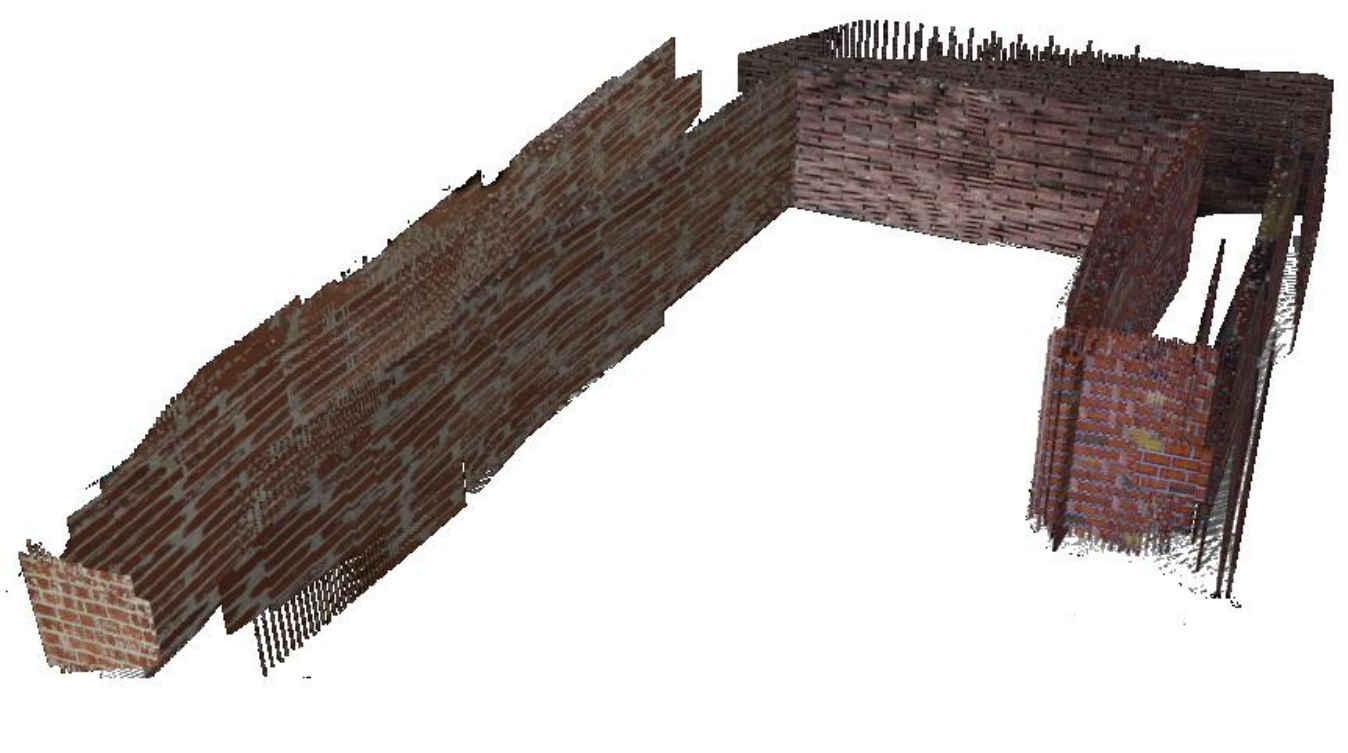}
	\label{subfig:small gamma husky 4.5 FBE model}
}
\hfil
\subfloat[Frontier based Exploration]{
	\includegraphics[width=\linewidth]{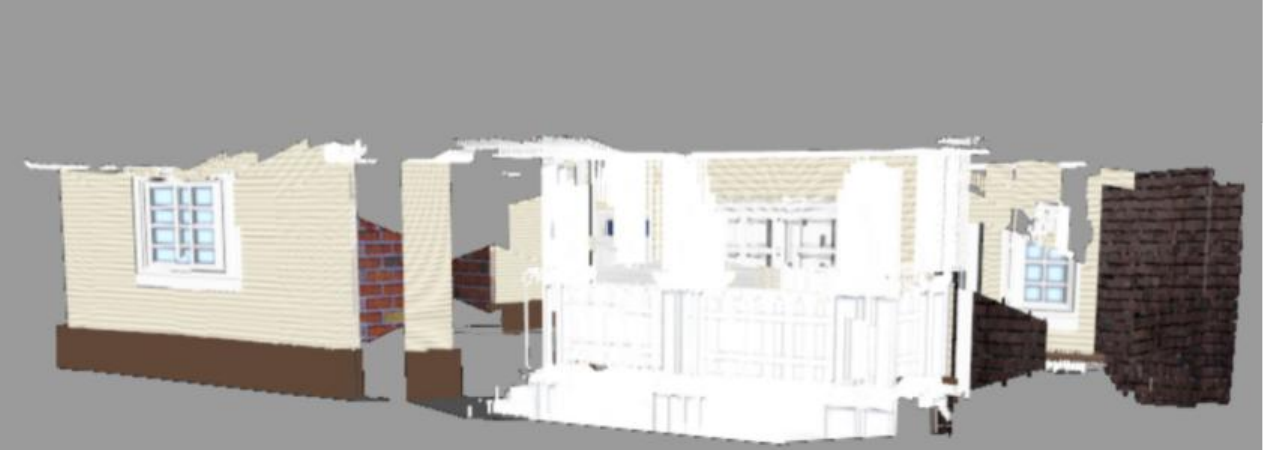}
	\label{subfig:house husky 4.5 FBE model}
}
\hfil
\subfloat[Model: Small $\Gamma$, Range: $4.5m$]{
	\includegraphics[width=1.5in]{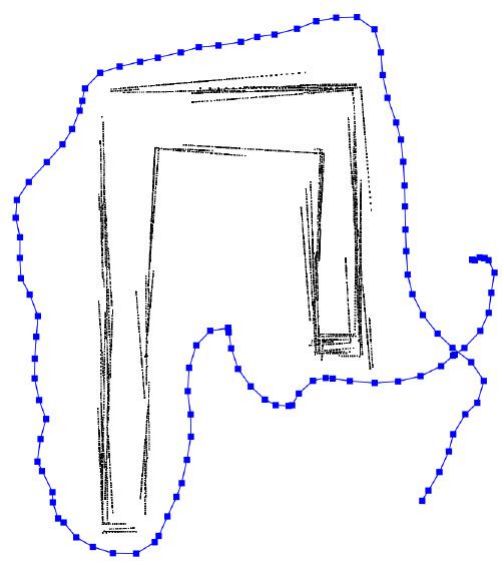}
	\label{subfig:small gamma husky 4.5 FBE}
}
\hfil
\subfloat[Model: House, Range: $4.5m$]{
	\includegraphics[width=1.5in]{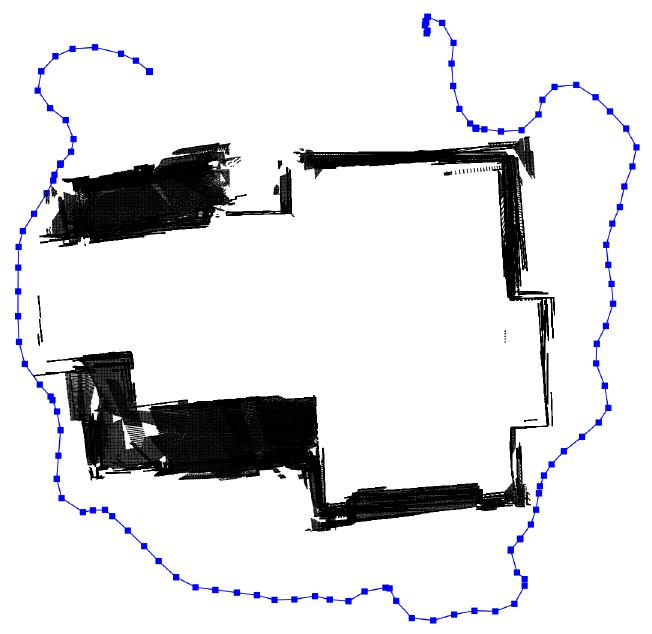}
	\label{subfig:house husky 4.5 FBE}
}
\hfil
\caption{(a-d) Comparison of the reconstructed model using our policies and FBE. 
Fig. (c) should be compared to Fig. \ref{subfig:reconstructed_model}.
(d,e) The trajectories prescribed by FBE for the Small $\Gamma$ and 
House structure are shown in blue.
}
\label{fig:simulation FBE}
\end{figure}

In order to get a quantitative measure of the structure coverage, 
we use again the CloudCompare software to compute essentially the projection
of the reconstructed model $\mathcal{C}$ on the reference point cloud $\mathcal{C}_R$. 
Namely, for each point in $\mathcal{C}$ we compute the closest point in $\mathcal{C}_R$.
Note that multiple points in $\mathcal{C}$ can have the 
same closest point in $\mathcal{C}_R$. In this case, we remove these 
duplicate points to obtain the \emph{unique closest point set}.
Then, as long as the reconstructed model aligns relatively well with the 
reference model, the cardinality of the unique closest point set 
is taken as our estimate of the structure coverage. 
Table \ref{tbl: simulation results 3} compares the level of structure
coverage achieved by our policies and FBE with a camera range of $4.5\text{m}$ 
for two of the environments considered.
For the Small $\Gamma$ model, our reference point cloud has $6,116$ points 
with a minimum distance of $0.1\text{m}$ between points. For the House model, 
our reference point cloud has $10,889$ points with a minimum distance
of $0.1\text{m}$ between points. Since the height of the House model is more 
than $H_{\max}$, we only take the portion of the reconstructed model up to 
the height $H_{\max}$ for computing the structure coverage and mean error 
for the two algorithms. 

Table \ref{tbl: simulation results 3} shows that 
our policies achieve a higher level of structure coverage than FBE 
for the environments considered and our proposed coverage metric.
Note also that the smooth trajectory prescribed by our policies is 
beneficial to the vSLAM module to achieve a better 
alignment and a lower value for the mean error in the 
reconstructed model, especially if the robot localization accuracy 
is poor.
A visual inspection of Fig. \ref{subfig:small gamma husky 4.5 model} 
and Fig. \ref{subfig:small gamma husky 4.5 FBE model} shows the improvement 
in model reconstruction when using our algorithms compared to FBE.

\section{Real-world Experiment}
\label{section: experiments}

\begin{figure}[thb]
\centerline{
	\subfloat{
		\includegraphics[width=0.43\linewidth]{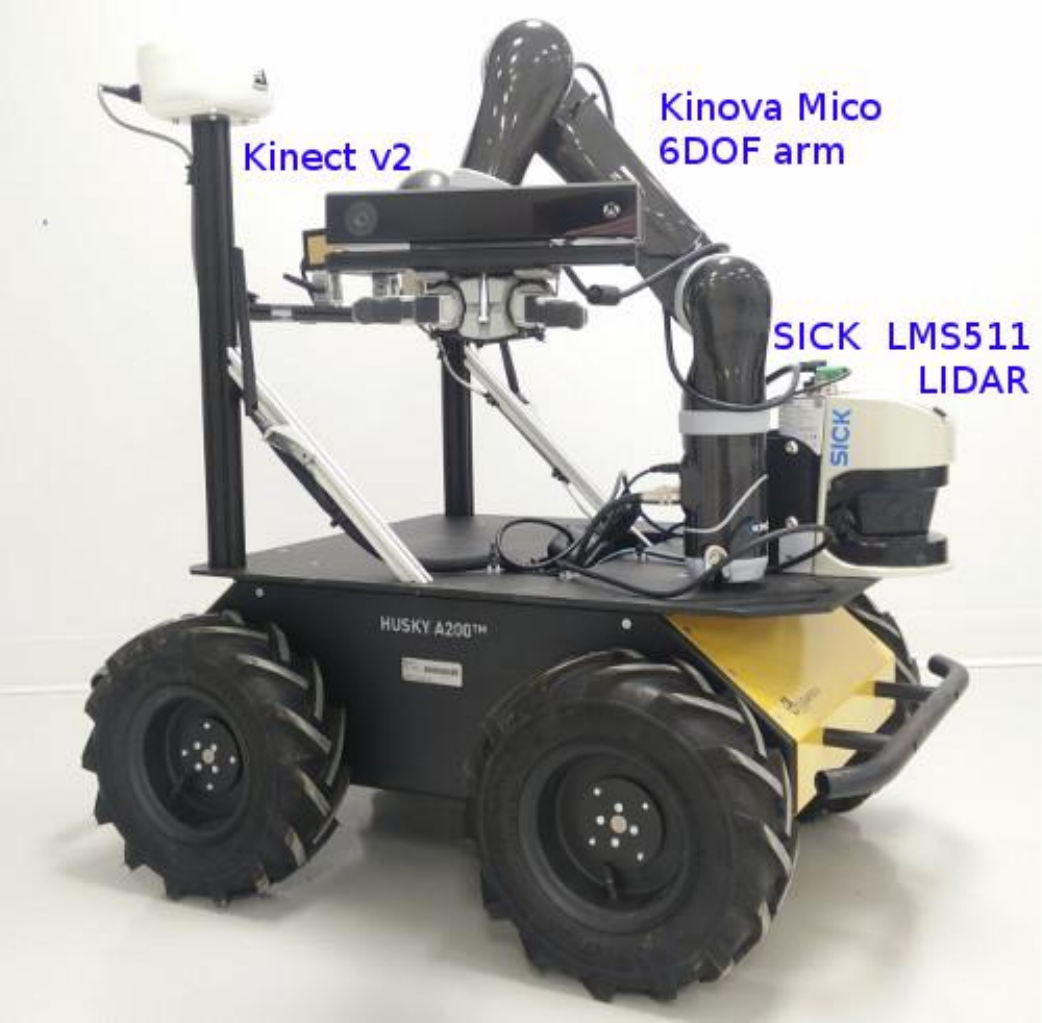}
        \label{fig:husky_exp_sensors}
    }%
    \subfloat{
		\includegraphics[width=0.57\linewidth]{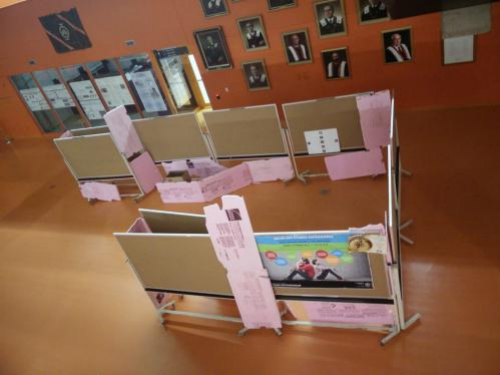}
        \label{fig:husky_exp_structure}
	}
}
	\centering
    \caption{
    Left: the Husky robot used in our experiments with the robotic arm, depth sensor and laser range scanner. 
    All other sensors seen are not used. 
    Right: the indoor structure being inspected.}
    \label{fig:husky_exp}
\end{figure}

Here we discuss an example of real-world experiment using the Husky robot shown in  
Fig. \ref{fig:husky_exp}, 
equipped with a Kinova robotic arm carrying a Kinect v2 depth sensor. 
A SICK laser range scanner is used only for obstacle detection. 
All computations are done in real-time on an embedded Intel i5 based computer  
without the help of a dedicated GPU. 
The structure was built from cork display panels and foam insulation boards. 
As shown in Fig. \ref{fig:husky_exp}, it 
mimics the small $\Gamma$ model used in simulation and its dimensions are $8.2\ \mathrm{m}$ by
$4\ \mathrm{m}$. To help the vSLAM algorithm detect a loop closure, 
we place a visual marker (a colorful poster seen on Fig. \ref{fig:husky_exp}) 
on the structure in front of the starting point of the robot.
Since the panels are similar on both sides we also put visual markers inside the structure 
to confirm that the cavity inspection has correctly mapped all of the inside. 
The most computationally intensive part of our algorithm is the detection of cavity entrances, 
see Section \ref{section: cavity entrance determination},
which takes a few seconds of computation for this structure at the end of the PE phase.

Due to the limited space available to maneuver around the structure, we reduce the obstacle sensing region
to a range of $1.8\ \mathrm{m}$ and to a 10 degree cone in front of the robot. The depth camera range 
is cut at $4.0\ \mathrm{m}$ and we set the desired wall distance $D$ to $1.5\ \mathrm{m}$. 
For odometry we use the extended Kalman filter (EKF) from \cite{MooreStouchGeneralizedEkf2014} 
with IMU and wheel odometry data as inputs. 
The output of the EKF is sent to RTAB-Map \cite{labbe14online} for mapping 
and localization purposes. 
Since we use a ground robot on flat terrain, 
we constrain RTAB-Map's mapping to 3 degrees of freedom (x, y and yaw angle).

\begin{figure}[h]
	\includegraphics[width=0.65\linewidth]{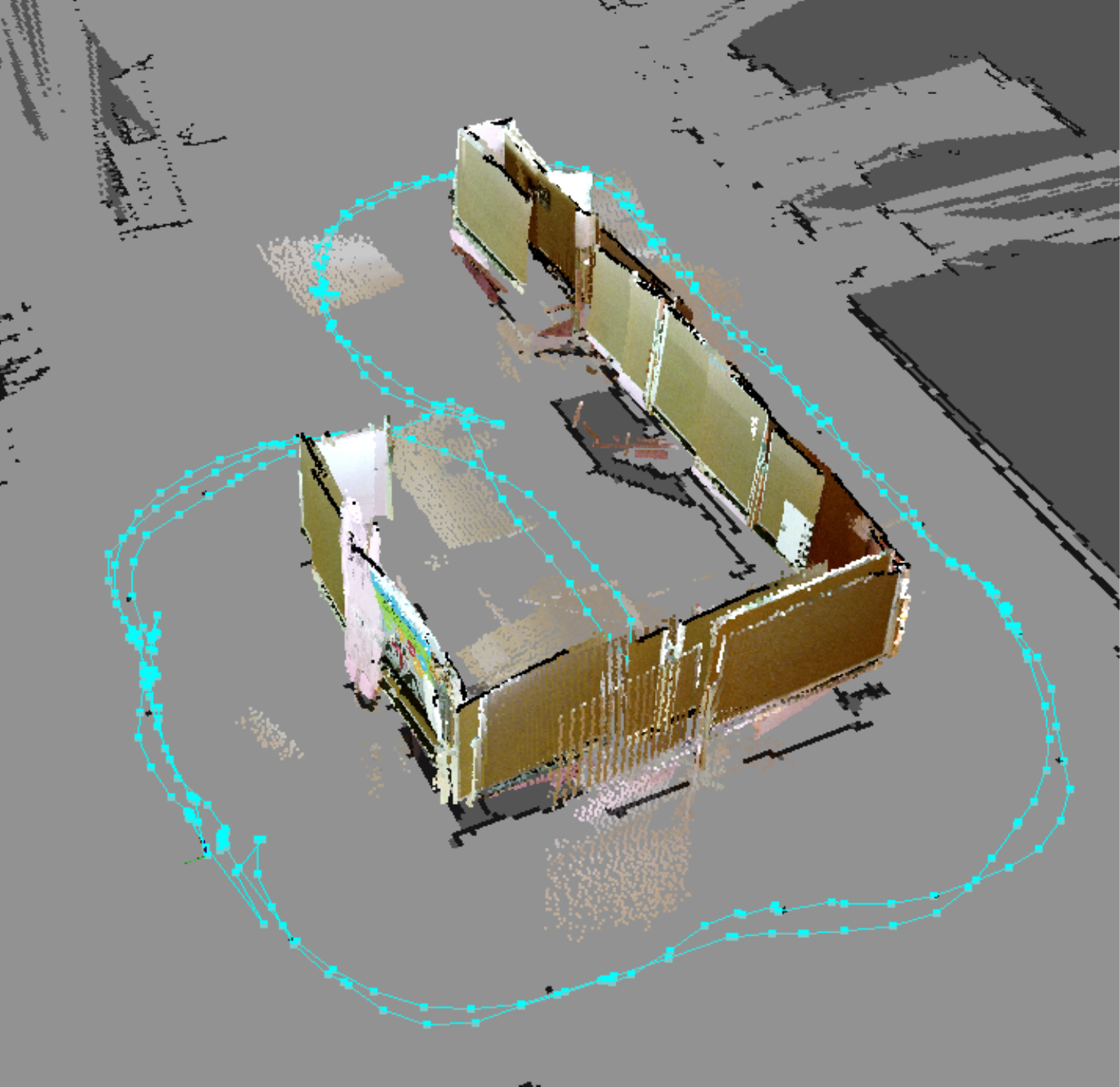}
	\centering
    \caption{An angled view of the constructed model. 
    The cyan lines with squares indicate the path taken by the robot.
    \label{fig:exp_angled_view}
    }
\end{figure}

The behavior of our algorithm is illustrated on the accompanying video, and the model
produced online with the vSLAM module is shown in Fig. \ref{fig:exp_angled_view}.
Overall, 
the executed trajectory confirms that our algorithm correctly performs 
perimeter exploration followed by cavity inspection. 
Noise in the depth measurements can be an issue if left unfiltered. 
However, proper calibration and applying a standard speckle and bilateral 
filter can alleviate the problem.   
In our experiments, we tuned the parameters of these filters by
placing the robot in a location where significant noise was measured.
We then increased the maximum speckle size and size of the bilateral filter 
window until most of the visible noise was removed. 

The experiment also illustrates some practical issues 
that can degrade mapping performance. First, because of the height 
of the boards used to build the structure and the limited space available
to navigate around it, the robotic arm was extended so that 
the depth sensor was at a height of $1.2 \mathrm{m}$. It then tended 
to shake during acceleration changes of the robot, 
making the sensor vulnerable to producing blurry images. 
This could be mitigated by a better control of the smoothness of the robot trajectories,
a stiffer orientable platform to hold the sensor, 
and by using a stereo camera with global shutters in bright daylight 
to reduce motion blur.
During testing we also noticed that the algorithm can be sensitive to gaps or "windows" 
in the structure. In front of a gap, the depth sensor can detect surfaces 
inside the structure, which can then perturb the goal calculation algorithm described 
in section \ref{section: next goal}. This can be addressed by reducing the range of
the sensor measurements to a value close to the desired distance $D$.

In summary, among the possible failure cases of our algorithm, we identified
during our experiments and simulations: 
i) waypoint determination errors due to the presence of gaps or windows in
the structure; and ii) incorrect global loop closures due to the possible
self-similarity of the structure, which can be mitigated by adding a unique
marker at the starting point.


\section{Conclusions} 
\label{section: conclusion}

This paper presents motion planning strategies that guide
a mobile ground robot carrying a camera or depth sensor to autonomously
explore the visible portion of a bounded three-dimensional 
structure. The proposed policies do not assume any prior
information about the size or geometry of the structure. 
Coupled with state-of-art vSLAM systems, our
strategies are able to achieve high coverage in the 
reconstructed model, given the physical limitations of the
platform. We illustrate the efficacy of our approach
via 3D simulations for different structure sizes, camera range 
and localization accuracy, and we have tested our system in real-world
experiments. In addition, a comparison of our policies
with the classical frontier based exploration algorithms clearly
shows the improvement in performance for a realistic structure such
as a house.


%




\ifCLASSOPTIONcaptionsoff
  \newpage
\fi



\bibliographystyle{IEEEtran}
\bibliography{IEEEabrv,biblio}
%
%
%

%
\begin{IEEEbiography}[{\includegraphics[width=1in,clip,keepaspectratio]{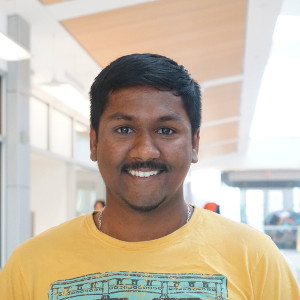}}]{Manikandasriram S.R.}
received the B.Tech and M.Tech degrees in electrical engineering 
from the Indian Institute of Technology Madras, India in 2016.
He is now working toward the Ph.D. degree in robotics at the Robotics Institute, 
University of Michigan, Ann Arbor, MI, USA. 
His research interests lie at the 
intersection of perception and control for autonomous vehicles. 
\end{IEEEbiography}

\begin{IEEEbiography}[{\includegraphics[width=1in,clip,keepaspectratio]{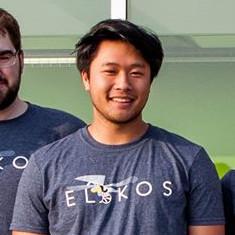}}]{Andr\'e Phu-Van Nguyen}
received his B.Eng. in Computer Engineering from Polytechnique Montreal in 2015 
and is currently pursuing his M.Sc.A. degree in electrical engineering at 
the same institution. 

His research interests revolve around making mobile robots smart and autonomous.
\end{IEEEbiography}

\begin{IEEEbiography}[{\includegraphics[width=1in,clip,keepaspectratio]{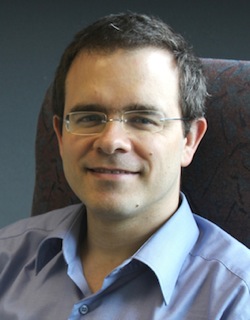}}]{Jerome Le Ny}
(S'05-M'09-SM'16) received the Engineering Degree from the \'Ecole Polytechnique, France, in 2001,
the M.Sc. degree in Electrical Engineering from the University of Michigan, Ann Arbor, in 2003, 
and the Ph.D. degree in Aeronautics and Astronautics from the Massachusetts Institute of Technology,
Cambridge, in 2008.
He is currently an Associate Professor with the Department of Electrical Engineering, 
Polytechnique Montreal, Canada, and a member of GERAD, a multi-university research 
center on decision analysis. From 2008 to 2012 he was a Postdoctoral Researcher with 
the GRASP Laboratory at the University of Pennsylvania. 
His research interests include planning under uncertainty,
robust and stochastic control, 
and networked control systems, 
with applications to 
autonomous multi-robot systems and intelligent infrastructures.
\end{IEEEbiography}








\vfill



\end{document}